\documentclass[sigconf]{acmart}
\AtBeginDocument{%
  }

\setcopyright{cc}
\setcctype{by}
\copyrightyear{2026}
\acmYear{2026}
\acmDOI{10.1145/3767308.3834932}
\acmConference[MM '26]{Proceedings of the 34th ACM International Conference on Multimedia}{November 10--14, 2026}{Rio de Janeiro, Brazil}
\acmBooktitle{Proceedings of the 34th ACM International Conference on Multimedia (MM '26), November 10--14, 2026, Rio de Janeiro, Brazil}
\acmISBN{979-8-4007-2213-4/2026/11}

\usepackage{balance}
\usepackage{multirow}
\usepackage{pifont}
\def\template{detailed long caption input template}
\def\encoder{dynamic caption encoder}
\def\decoder{text-adaptive decoder}
\def\eg{\textit{e.g.}}
\def\ie{\textit{i.e.}}

\begin{document}

\let\originalcite\cite
\renewcommand{\cite}[1]{%
  \textcolor{ACMPurple}{\originalcite{#1}}%
}

\let\originalref\ref
\renewcommand{\ref}[1]{%
  \textcolor{ACMPurple}{\originalref{#1}}%
}

\title[Guiding Robust Monocular Depth Estimation in Challenging Scenarios via Detailed Long Captions]{Beyond Visual Ambiguity: Guiding Robust Monocular Depth Estimation in Challenging Scenarios via Detailed Long Captions}


\author{Junrui Zhang}
\orcid{0009-0004-8018-0458}
\affiliation{
  \institution{School of Artificial Intelligence and Automation, Huazhong University of Science and Technology}
  \city{Wuhan}
  \country{China}
}
\email{junrui@hust.edu.cn}

\author{Jiaqi Li}
\orcid{0009-0004-7799-3407}
\affiliation{%
  \institution{School of Artificial Intelligence and Automation, Huazhong University of Science and Technology}
  \city{Wuhan}
  \country{China}
}
\email{lijiaqi\_mail@hust.edu.cn}

\author{Yiran Wang}
\orcid{0000-0002-2785-9638}
\affiliation{%
  \institution{School of Artificial Intelligence and Automation, Huazhong University of Science and Technology}
  \city{Wuhan}
  \country{China}
}
\email{wangyiran@hust.edu.cn}

\author{Liao Shen}
\orcid{0000-0002-2423-4835}
\affiliation{%
  \institution{School of Artificial Intelligence and Automation, Huazhong University of Science and Technology}
  \city{Wuhan}
  \country{China}
}
\email{leoshen@hust.edu.cn}

\author{Zhiguo Cao}
\correspondingauthor
\orcid{0000-0002-9223-1863}
\affiliation{
  \institution{School of Artificial Intelligence and Automation, Huazhong University of Science and Technology}
  \city{Wuhan}
  \country{China}
}
\email{zgcao@hust.edu.cn}

\renewcommand{\shortauthors}{Junrui Zhang, Jiaqi Li, Yiran Wang, Liao Shen and Zhiguo Cao}

\begin{abstract}
Monocular depth estimation (MDE) faces challenges with non-Lambertian surfaces and adverse weather conditions due to the visual ambiguities inherent in single-image limited information. Existing works address them in isolation via image inpainting or augmentation, yielding limited robustness gains. Language, as a powerful complementary modality to vision, is demonstrated to enhance the visual perception capabilities of vision-language models (VLMs) via detailed long captions. However, prior language-integrated MDE methods fail to fully harness this potential due to short text input with limited information, coarse global text feature learning, and limited language guidance during depth decoding. To address these limitations, we propose CapDepth, a novel framework for robust MDE that leverages guidance from detailed long captions to alleviate visual ambiguities in both challenging scenarios. First, we design a detailed long caption input template that explicitly conveys rich spatial relationships among multiple atom sentences. Second, a dynamic caption encoder is introduced to extract fine-grained depth-relevant text features via progressive masked attention. Finally, we propose a text-adaptive decoder that guides enhanced depth decoding with text features via stable adaptive layer normalization. Extensive experiments validate the efficacy of CapDepth, which outperforms state-of-the-art methods, achieving depth error reductions of 25.0\% on non-Lambertian surfaces and 22.0\% under adverse weather conditions.
\end{abstract}

\ccsdesc[500]{Computing methodologies~Scene understanding}

\keywords{Depth Estimation; Challenging Scenarios; Language Guidance}
\begin{teaserfigure}
  \includegraphics[width=\textwidth]{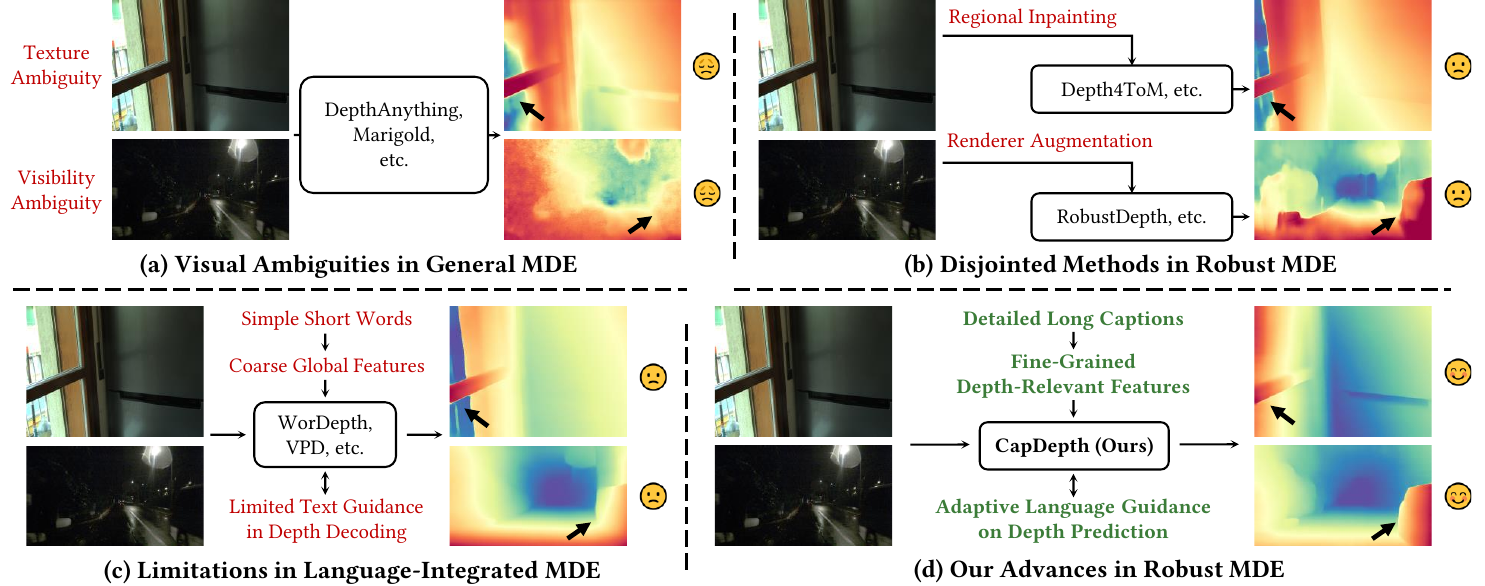}
  \vspace{-20pt}
  \caption{
  Comparisons of different paradigms for robust monocular depth estimation (MDE).
  (a) General MDE methods \cite{da,metric3d,marigold,depthfm,lotus} struggle with non-Lambertian surfaces (texture ambiguity) and adverse weather conditions (visibility ambiguity).
  (b) Scenario-specific methods \cite{depth4tom,robustdepth} address them in isolation with limited robustness improvement.
  (c) Prior language-integrated methods \cite{wordepth,vpd} fail to fully harness the language modality.
  (d) We propose CapDepth that leverages detailed long captions to provide comprehensive guidance, achieving significant improvement across various challenging scenarios.
  }
  \label{fig:teaser}
\end{teaserfigure}


\maketitle

\section{Introduction}
\label{sec:intro}

Monocular depth estimation (MDE) aims to predict a dense depth map from a single image, which is critical for applications like novel view synthesis \cite{nvs}, bokeh rendering \cite{BokehMe}, and virtual reality \cite{vr}.
However, its acknowledged ill-posed and under-constrained nature makes MDE susceptible to the visual ambiguities in challenging scenarios \cite{tricky2024,ntire2024,depth4tom,robustdepth,eccvw}.
In particular, non-Lambertian surfaces (\eg, glass or mirror) introduce the texture ambiguity \cite{booster}, while adverse weather conditions (\eg, rain or night) cause the visibility ambiguity \cite{nuscenes}.
While general MDE methods \cite{da,da2,metric3d,metric3d2,marigold,depthfm,lotus}, such as Depth Anything \cite{da}, achieve impressive MDE performance, they still struggle with the visual ambiguities, leading to degraded results in both challenging scenarios as illustrated in Fig.~\ref{fig:teaser}~(a).

To improve robustness in challenging scenarios, some disjointed methods \cite{robustdepth,depth4tom} rely on scenario-specific data augmentation strategies for robust MDE as shown in Fig.~\ref{fig:teaser}~(b).
For example, Depth4ToM \cite{depth4tom} employs the semantic segmentation model \cite{mirrornet} to segment transparent or mirror (ToM) surfaces and inpaints them with uniform colors.
RobustDepth \cite{robustdepth} uses a physics-based renderer to synthesize rain and fog augmentations in driving scenes.
Despite marginal improvements, their scenario-specific nature makes them inapplicable to other challenging scenarios.
This naturally raises an intriguing question: \textit{Is it possible to develop a unified framework that simultaneously alleviates the visual ambiguities arising from both non-Lambertian surfaces and adverse weather conditions?}

To answer this question, we explore language, a powerful complementary modality that provides explicit guidance for scene understanding \cite{sharegpt4v,vlm1,vlm2,vlm3,vlm4}.
Recent advances in vision-language models (VLMs) \cite{sharegpt4v,longclip} further demonstrate that detailed long captions, rather than simple short texts, significantly enhance visual perception capabilities in tasks such as object localization and segmentation \cite{sharegpt4v,vlm1,vlm2,vlm3,vlm4}.
This suggests that detailed long captions could potentially guide MDE in alleviating visual ambiguities in challenging scenarios, achieving enhanced robustness.
For example, in the nighttime driving scene (Fig.~\ref{fig:teaser}), a detailed long caption input containing a sentence like ``A vehicle is in front of and to the right of the trees.'' provides richer spatial clues than a brief text input like ``a photo of a car'' \cite{vpd}.
Such language guidance could effectively alleviate visibility ambiguities, enabling robust MDE both on non-Lambertian surfaces and under adverse weather conditions.

To harness such rich language guidance, three key questions naturally arise:
(1) What formal template should the input text follow?
(2) How to extract effective text features to guide MDE?
(3) How can text features effectively guide depth prediction?
Prior language-integrated MDE methods \cite{vpd,wordepth} provide suboptimal solutions.
Specifically, they employ simple short text input templates with limited information (\eg, ``a photo of a <class name>'' \cite{vpd}).
Moreover, they extract coarse global text features (\eg, the [CLS] token \cite{vpd,wordepth}) and rely on features that exhibit a gap from the precise geometric reasoning required for MDE \cite{evp} (\eg, U-Net noise prediction features \cite{vpd}) for depth decoding without language guidance.
Consequently, they fail to fully exploit the potential of language to alleviate visual ambiguities as shown in Fig.~\ref{fig:teaser}~(c).

To better address the three questions, we propose CapDepth (Fig.~\ref{fig:teaser}~(d)), a robust MDE framework that leverages detailed long captions to guide the alleviation of visual ambiguities in challenging scenarios (non-Lambertian surfaces and adverse weather conditions).
Correspondingly, CapDepth consists of three key components:
(1) a \template~composed of spatially descriptive sentences, which explicitly conveys rich spatial information to guide MDE,
(2) a \encoder~that captures fine-grained depth-relevant text features via progressive masked attention,
(3) a \decoder~that leverages stable adaptive layer normalization to bridge the aforementioned feature gap, guiding enhanced depth decoding with text features.

We conduct extensive experiments on established benchmarks, including Booster \cite{booster}, ClearGrasp \cite{cleargrasp}, nuScenes \cite{nuscenes}, and DrivingStereo \cite{drivingstereo}.
Comprehensive comparisons are performed against prevalent general MDE methods \cite{da,da2,metric3d,metric3d2,marigold,lotus,depthfm}, prior robust MDE approaches \cite{robustdepth,depth4tom}, and existing language-integrated MDE models \cite{vpd,wordepth}.
Results demonstrate that CapDepth effectively leverages guidance from detailed long captions to alleviate visual ambiguities arising from both non-Lambertian surfaces and adverse weather conditions.
Consequently, CapDepth significantly improves MDE robustness in challenging scenarios, achieving depth error reductions of 25.0\% and 22.0\% in the respective scenarios.

Our main contributions can be summarized as follows:
\begin{itemize}
\item A \template~that explicitly conveys rich spatial relationships via multiple atom sentences to provide language guidance.
\item A \encoder~that extracts fine-grained depth-relevant text features from detailed long captions via progressive masked attention.
\item A \decoder~that leverages stable adaptive layer normalization to guide the enhanced depth decoding process with the extracted text features.
\end{itemize}

\begin{figure*}[t]
  \centering
  \includegraphics[width=\linewidth]{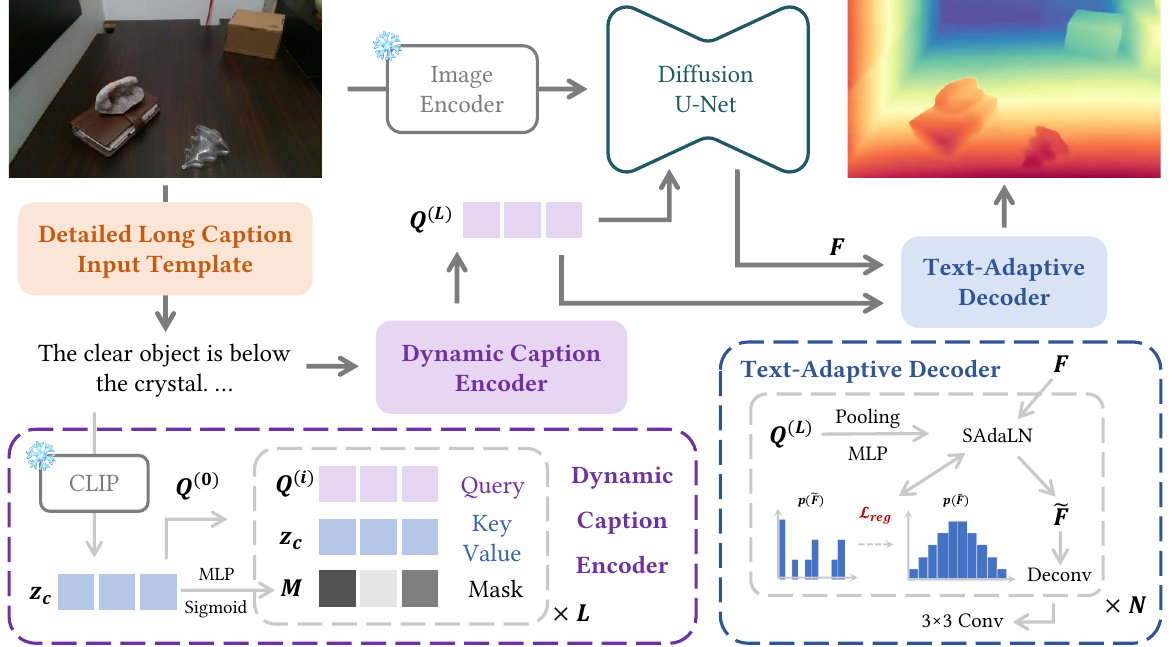}
  \vspace{-15pt}
  \caption{Overview of CapDepth. The detailed long caption input template comprises multiple basic atom sentences, each describing the spatial relationships between a pair of objects. Subsequently, the dynamic caption encoder with progressive masked querying blocks extracts fine-grained depth-relevant text features. Finally, the text-adaptive decoder leverages these text features to guide the depth decoding process via stable adaptive layer normalization (SAdaLN).}
  \vspace{-10pt}
  \label{fig:method}
\end{figure*}

\section{Related Work}
\label{sec:related}

\subsection{General MDE}
Prevalent general monocular depth estimation (MDE) works broadly fall into two paradigms: generative methods~\cite{marigold,depthfm,ch3depth} using diffusion \cite{sd} priors, and discriminative models~\cite{da,da2,metric3d,metric3d2} trained on massive data.
Despite impressive results, they exhibit degraded performance across non-Lambertian surfaces and adverse weather conditions.
This stems from the scarcity of accurate depth annotations in such scenarios, which are crucial for these methods to alleviate visual ambiguities.
Such data deficiency is fundamentally caused by the inherent limitations of active sensors (\eg, LiDAR and structured light).
To bypass this data bottleneck, we propose leveraging language modality.
By harnessing rich spatial information in detailed long captions, our method guides MDE to alleviate visual ambiguities, enhancing robustness in these challenging scenarios.

\subsection{Robust MDE}
To achieve robust MDE on non-Lambertian surfaces and under adverse weather conditions, recent efforts \cite{robustdepth,depth4tom} address them independently.
Specifically, Depth4ToM \cite{depth4tom} inpaints segmented transparent or mirror (ToM) surfaces with uniform colors, while RobustDepth \cite{robustdepth} synthesizes weather effects via physics-based rendering.
However, such augmentations introduce a domain gap between synthesized images and the real world, and weather augmentation apparently cannot handle non-Lambertian surfaces, and vice versa.
In contrast, our CapDepth leverages detailed long captions as a unified guidance to simultaneously alleviate visual ambiguities and enhance robustness across both challenging scenarios.

\subsection{Language-Integrated MDE}
Recent MDE studies explore incorporating text \cite{vpd,wordepth}.
For instance, extracting [CLS] token from simple short text, VPD \cite{vpd} uses noise prediction features for depth decoding, while WorDepth \cite{wordepth} uses conditional sampling features.
However, they exhibit three limitations:
(1) simple short text input with limited information;
(2) coarse global text feature extraction;
(3) relying on features that exhibit a gap from precise geometric reasoning required for MDE \cite{evp}.
In contrast, our CapDepth addresses these with: 
(1) a \template~providing rich spatial clues;
(2) a \encoder~extracting fine-grained depth-relevant text features;
(3) a \decoder~leveraging text features to bridge the feature gap and guide enhanced depth decoding.

\section{Method}
\label{sec:method}

In this section, we first provide an overview in Sec.~\ref{sec:overview}.
Then we elaborate on the \template~in Sec. \ref{sec:dlc}.
The \encoder~will be discussed in Sec.~\ref{sec:encoder}.
The \decoder~is described in Sec.~\ref{sec:decoder}.

\subsection{Overview}
\label{sec:overview}
Here we present an overview of our CapDepth framework.
Given a single image $I\in \mathbb{R}^{H\times W\times 3}$ and its corresponding caption $C$ as inputs, our goal is to predict a dense depth map $D\in\mathbb{R}^{H\times W}$.
$H$ and $W$ represent the height and width of the image.
As discussed in Sec.~\ref{sec:intro}, the keys to the exploitation of language can be divided into three parts: (1) text input template; (2) text feature extraction; and (3) text guidance on depth decoding.
Accordingly, as depicted in Fig.~\ref{fig:method}, CapDepth proposes (1) a \template~that provides rich spatial information; (2) a \encoder~for fine-grained depth-relevant text feature learning; and (3) a \decoder~for an enhanced depth decoding.

First, as discussed in Sec.~\ref{sec:intro}, to guide MDE in challenging scenarios with visual ambiguities, the language modality needs to convey rich spatial information.
To this end, we propose a \template, which consists of many structured atom sentences that concisely convey such spatial information.
Subsequent analysis and experiments confirm the efficacy of this template over previous simple short text input templates \cite{vpd,wordepth}.

Second, although the input caption contains rich spatial information, extracting coarse global features (\eg, the [CLS] token) as in prior works \cite{vpd,wordepth} inevitably discards the valuable guidance provided by the language modality.
To address this, we introduce a \encoder~that learns to extract fine-grained text features via progressive masked querying blocks.
Concurrently, a learnable soft mask is employed to dynamically distinguish depth-relevant tokens for guiding robust MDE.

Finally, text features and image features first enter the diffusion U-Net \cite{sd} for initial interaction, obtaining the internal U-Net representations.
However, these representations are originally optimized for noise prediction \cite{sd}, which creates a gap between the denoising objective and the precise geometric reasoning required for MDE \cite{evp}.
To bridge this gap and fully exploit language guidance during depth decoding, we propose a \decoder.
Specifically, it leverages the text features to modulate the visual features via stable adaptive layer normalization (SAdaLN).
By explicitly injecting these text-based spatial clues into the depth decoding process, our decoder effectively aligns the visual features with the MDE task, thereby providing strong guidance for more accurate and robust depth predictions in challenging scenarios.

\begin{table}[b]
\vspace{-10pt}
\caption{
    Linguistic statistics of spatial information across text inputs of various methods.
    ``Spatial Word Ratio'' denotes the proportion of spatial relationship words per input text.
    ``Spatial-to-object Ratio'' denotes the ratio of spatial relationship words to object-related words.
    ``Average Word Count'' denotes the average number of words per input text.
}
\label{tab:dlc}
\vspace{-10pt}
\setlength{\tabcolsep}{3pt}
\begin{tabular}{cccc}
    \toprule
\multirow{2}{*}{Text Input} & Spatial Word  & Spatial-to-object  &Average \\
 & Ratio $\uparrow$ & Ratio $\uparrow$  &Word Count $\uparrow$ \\
    \midrule
VPD~\cite{vpd}& 0.02& 0.12&5.30\\
WorDepth~\cite{wordepth}& 0.09&0.36&8.86\\
Ours&\textbf{0.34}&\textbf{3.06}&\textbf{45.25}\\
    \bottomrule
\end{tabular}
\end{table}

\subsection{Detailed Long Caption Input Template}
\label{sec:dlc}
As discussed in Sec.~\ref{sec:intro}, recent works \cite{sharegpt4v,longclip} demonstrate that detailed long captions, rather than simple short texts (\eg, ``a photo of a <class name>'' \cite{vpd}), can significantly enhance the visual perception capabilities of vision-language models (VLMs).
To this end, we aim to design a \template, enabling the input text to better guide robust MDE.
Specifically, to be tailored for MDE, the input text should explicitly convey the spatial relationships among objects in the scene to improve scene perception.
To achieve this, we first formulate an atom sentence template: ``$O_A$ \{spatial relationship phrases\} $O_B$.'', where $O_A$ and $O_B$ represent the specific objects (\eg, ``red apples'', ``black vehicles'', etc.), and the spatial relationship phrases can be any commonly used spatial expressions (\eg, ``be to the left of'', ``be in front of'', etc.).
Building upon this, our \template~is constructed by seamlessly stacking such atom sentence templates.
Ultimately, by prompting human annotators or VLMs to generate texts that strictly adhere to this template, we obtain the desired detailed long caption input, which contains rich spatial information about the scene.
In practice, we employ VLMs \cite{internvl} to generate the detailed long captions.
Additional details and concrete examples of the detailed long captions are provided in the supplementary material.

To validate the effectiveness of our proposed \template, we conduct comparative experiments against various simple short text input templates used in previous language-integrated MDE methods \cite{vpd,wordepth}, as shown in Tab.~\ref{tab:dlc} and Tab.~\ref{tab:ab_text}.
For instance, the simple short text input template of VPD \cite{vpd} is ``a photo of a <class name>''.
And an example for WorDepth \cite{wordepth} is ``A bedroom with a bed and a table.'', which lacks precise rich spatial information as well.
We provide an analysis of Tab.~\ref{tab:dlc} here (see supplementary material for experimental details) and reserve the discussion of Tab.~\ref{tab:ab_text} for Sec.~\ref{sec:template}.
The linguistic statistics in Tab.~\ref{tab:dlc} demonstrate that the text input following our designed \template~exhibits a significantly higher density of spatial relationship words (34\% vs. 9\%), averaging over 3 spatial relationship words per object compared to merely 0.36 in the prior work \cite{wordepth}, alongside substantially longer sequence lengths (45.25 vs. 8.86 words).
This confirms that our proposed \template~could provide richer spatial information compared to existing language-integrated MDE methods.

\subsection{Dynamic Caption Encoder}
\label{sec:encoder}

Although the detailed long caption input contains rich spatial information, extracting coarse global features (\eg, the [CLS] token) as in prior works \cite{vpd,wordepth} inevitably causes information loss, leading to suboptimal language guidance for robust MDE (see analysis of Tab.~\ref{tab:ab_text} in Sec.~\ref{sec:template}).
To address this, we introduce a \encoder~which features progressive masked querying blocks to extract fine-grained depth-relevant linguistic representations.
Specifically, as illustrated in Fig.~\ref{fig:method}, given the detailed long caption input $C$, we first extract its CLIP \cite{clip,longclip} features $z_C\in\mathbb{R}^{n\times d_C}$, where $n$ denotes the number of tokens and $d_C$ represents the feature dimension of each token.
However, not all tokens provide meaningful spatial information equally (\eg, the model should pay more attention to tokens like ``behind'' over functional words like ``the'').
To address this, we leverage a lightweight network consisting of an MLP followed by a sigmoid activation to predict the relevance scores based on $z_C$ for the $n$ tokens.
These scores are then expanded into an $n\times n$ soft mask $M$ by replicating the scores across $n$ rows.
Subsequently, progressive masked querying blocks are proposed to extract fine-grained text features from $z_C$ based on $M$ to guide MDE.
These blocks enable query interactions with $z_C$, focusing on depth-relevant words while suppressing less relevant tokens through masked attention.
Formally, we first initialize the query $Q^{(0)}$ as $z_C$.
For the $i$-th block ($i$ from $1$ to $L$), it takes $z_C$ and $Q^{(i-1)}$ as input and produces $Q^{(i)}$ according to Eq.~\ref{eq:block}:
\begin{equation}
\label{eq:block}
Q^{(i)}=softmax(\frac{Q^{(i-1)}W_Q(z_CW_K)^T\odot M}{\sqrt{d_k}})z_CW_V,
\end{equation}
where $W_Q,W_K,W_V\in \mathbb{R}^{d_C\times d_k}$ are learnable projection matrices, $d_k$ denotes the dimension of hidden states, and $\odot$ denotes the element-wise product.
After $L$ progressive masked querying blocks, we obtain the final fine-grained depth-relevant text features $Q^{(L)}$.

\begin{figure}[!t]
  \centering
  \includegraphics[width=\linewidth]{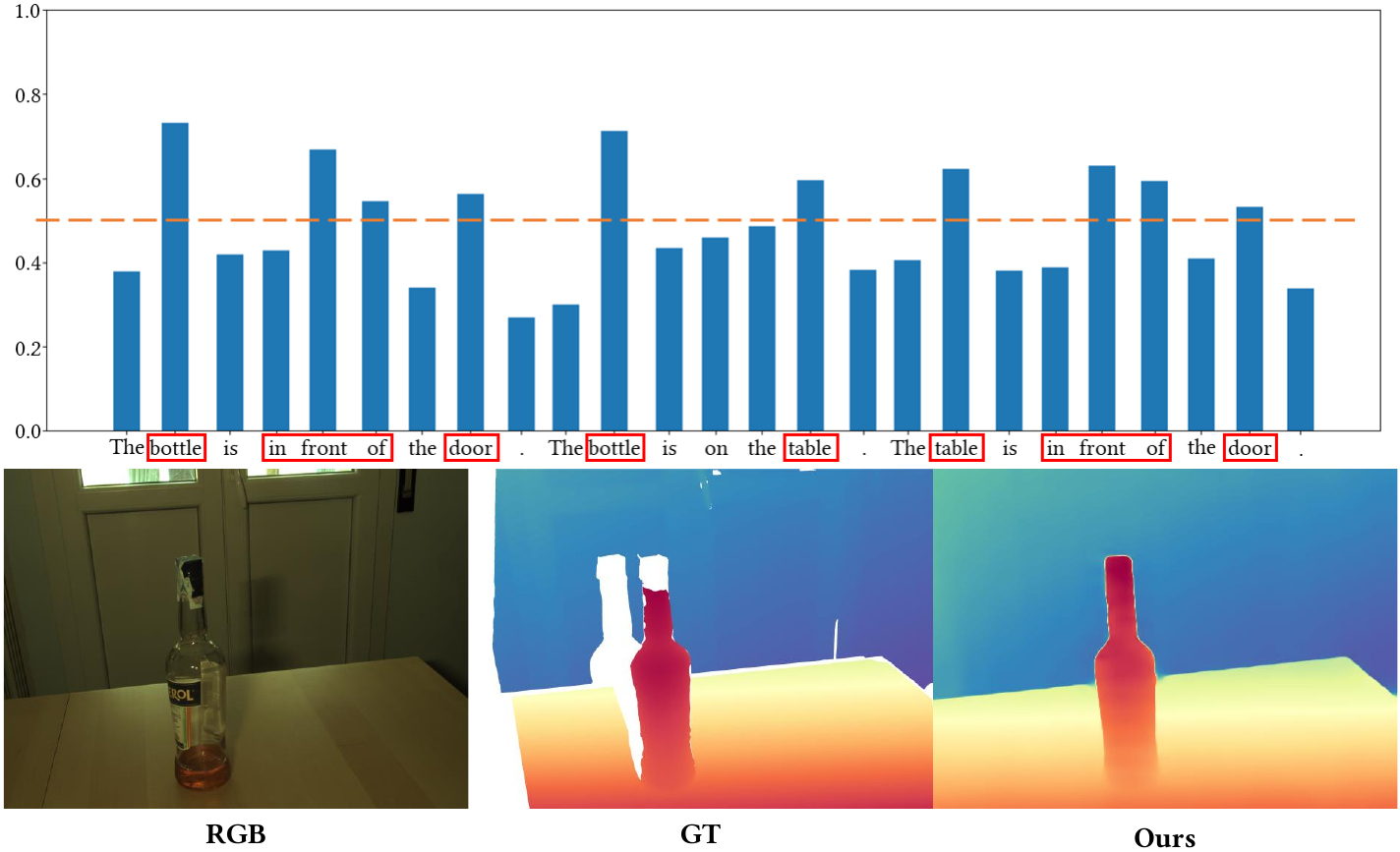}
  \vspace{-22pt}
  \caption{
  Visualization of relevance scores for input tokens.
  }
  \vspace{-18pt}
  \label{fig:mask}
\end{figure}

To demonstrate the effectiveness of our proposed \encoder, we conduct extensive experiments, including ablation studies in Tab.~\ref{tab:ab} of Sec.~\ref{sec:ab_main}, additional analysis in the supplementary material, and the visualization of the scores in the mask $M$ in Fig.~\ref{fig:mask}.
Here we focus on analyzing Fig.~\ref{fig:mask} and defer the remaining discussions to the corresponding sections.
As illustrated in Fig.~\ref{fig:mask}, our proposed \encoder~adaptively captures object words (\eg, ``bottle'', ``door'') and spatial relationship phrases (\eg, ``in front of'') from the input text, assigning them higher weights than other less informative tokens (\eg, the token for the period ``.''), effectively guiding robust MDE to produce results that closely approximate the ground truth (GT).

\subsection{Text-Adaptive Decoder}
\label{sec:decoder}
As illustrated in Fig.~\ref{fig:method}, following our baseline \cite{vpd}, the input image $I$ is first processed by a frozen pre-trained variational autoencoder (VAE) \cite{vae,sd} to obtain its latent representation.
Subsequently, text features $Q^{(L)}$ and these image latents are fed into a pre-trained diffusion U-Net \cite{sd} with the timestep set to $t=0$ \cite{vpd}.
We extract the hierarchical pyramid features from the up-sampling layers of the diffusion U-Net \cite{vpd}.
These features are then up-sampled to a uniform resolution and concatenated along the channel dimension \cite{vpd}, yielding the final U-Net image features, denoted $F$.
However, as discussed in Sec.~\ref{sec:overview}, the feature $F$ inherently represents noise prediction features, which exhibit a gap from the precise geometric reasoning required for MDE \cite{evp}.
To address this gap, we propose a \decoder~that incorporates language guidance to enhance the depth decoding of $F$ via stable adaptive layer normalization (SAdaLN).
Specifically, as shown in Fig.~\ref{fig:method}, text features $Q^{(L)}$ are first average-pooled and fed into a lightweight MLP to produce two vectors, $a$ and $b$, which share the same dimension as $F$.
These vectors from $Q^{(L)}$ are fused with $F$ via Eq.~\ref{eq:sadaln}:
\begin{equation}
\label{eq:sadaln}
\tilde{F} =\mathrm{LayerNorm}(F) \cdot a + b,
\end{equation}
where $\mathrm{LayerNorm}$ denotes the layer normalization.
The updated features $\tilde{F}$ are then up-sampled through a $2\times2$ de-convolution module.
This fusion and up-sampling process is repeated $N$ times, obtaining the final feature representation, which will be regressed to the depth map $D$ via a lightweight $3\times3$ convolution head.

\begin{figure}[!t]
  \centering
  \includegraphics[width=\linewidth]{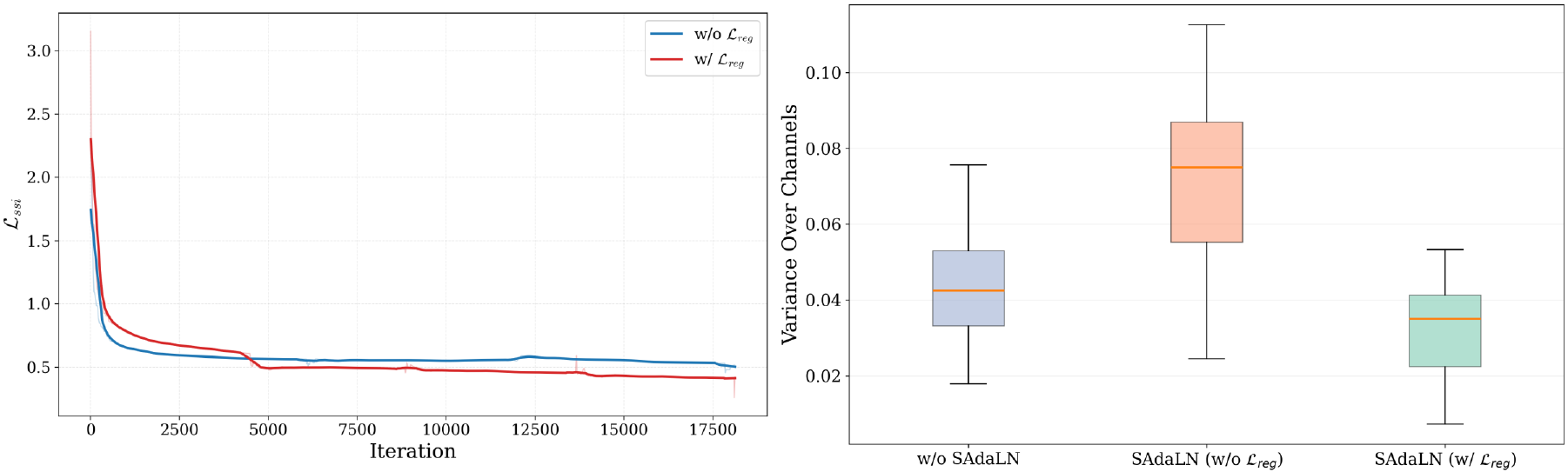}
  \vspace{-20pt}
  \caption{
  The training loss curves of $\mathcal{L}_{ssi}$ (left).
  The box plot of the channel-wise variance of the feature $\tilde{F}$ (right).
  }
  \vspace{-15pt}
  \label{fig:loss}
\end{figure}

The model can then be trained using the standard MDE loss $\mathcal{L}_{ssi}$ \cite{midas}, following previous methods \cite{da,da2}.
However, optimizing with only $\mathcal{L}_{ssi}$ hinders further convergence during training.
As illustrated in Fig.~\ref{fig:loss}~(left), the training loss experiences a prolonged plateau between 2500 and 15000 iterations.
To investigate this issue, we analyze the distribution of $\tilde{F}$.
As shown in Fig.~\ref{fig:loss}~(right), applying Eq.~\ref{eq:sadaln} without any constraints significantly enlarges the variance of $\tilde{F}$ along the channel dimension compared to the baseline without Eq.~\ref{eq:sadaln}.
This primarily stems from the substantial variations among different input texts, which cause fluctuations in the learned $a$ and $b$, thereby making it difficult for the model to learn the distribution for $\tilde{F}$.
To alleviate the learning difficulty and facilitate better convergence, we explicitly constrain the distribution of $\tilde{F}$ to follow a standard normal distribution, \ie, $\tilde{F}\sim \mathcal{N}(0,I)$.
Given that $\tilde{F}\sim \mathcal{N}(b,a^2 I)$ according to Eq.~\ref{eq:sadaln}, we minimize the Kullback-Leibler (KL) divergence between $\mathcal{N}(b,a^2 I)$ and $\mathcal{N}(0,I)$ via the derived Eq.~\ref{eq:kl} (please refer to the supplementary material for detailed derivations):
\begin{equation}
\label{eq:kl}
\mathcal{L}_{reg}=-\log |a|+\frac{a^2+b^2}{2} -\frac{1}{2}.
\end{equation}
Consequently, the overall optimization objective is formulated as:
\begin{equation}
\mathcal{L}=\mathcal{L}_{ssi}+\lambda\mathcal{L}_{reg},
\label{eq:loss}
\end{equation}
where $\lambda$ is a hyperparameter that balances the MDE loss and the regularization term.
By incorporating $\mathcal{L}_{reg}$, as depicted in Fig.~\ref{fig:loss}~(right), the channel-wise variance of $\tilde{F}$ is significantly reduced.
This effectively alleviates the learning burden on the model, thereby facilitating an improved convergence, as evidenced in Fig.~\ref{fig:loss}~(left).
By integrating Eq.~\ref{eq:sadaln} and Eq.~\ref{eq:kl}, SAdaLN effectively leverages text features to guide depth decoding, achieving enhanced robust MDE.

\begin{table*}[!t]
\caption{
Zero-shot comparisons in robust MDE.
DA denotes Depth Anything.
``---'' means that Metric3D models are trained on DrivingStereo (violating the zero-shot setting), and Depth4ToM and RobustDepth are designed for non-Lambertian surfaces and adverse weather conditions, respectively, making them inapplicable to the other challenging scenario.
}
\vspace{-10pt}
\label{tab:main}
    \setlength{\tabcolsep}{2.0pt}
    \begin{tabular}{ccccccccccccccccccccccccc}
    \toprule
\multirow{3}{*}{Method} & \multicolumn{4}{c}{Booster \cite{booster}} & \multicolumn{4}{c}{ClearGrasp \cite{cleargrasp}} & \multicolumn{2}{c}{nuScenes \cite{nuscenes}} & \multicolumn{6}{c}{DrivingStereo \cite{drivingstereo}}\\
& \multicolumn{2}{c}{ToM} & \multicolumn{2}{c}{All} & \multicolumn{2}{c}{ToM} & \multicolumn{2}{c}{All} & \multicolumn{2}{c}{night-rain} & \multicolumn{2}{c}{cloudy} & \multicolumn{2}{c}{foggy} & \multicolumn{2}{c}{rainy} \\
& AbsRel$\downarrow$ & $\delta_1\uparrow$ & AbsRel$\downarrow$ & $\delta_1\uparrow$ & AbsRel$\downarrow$ & $\delta_1\uparrow$ & AbsRel$\downarrow$ & $\delta_1\uparrow$& AbsRel$\downarrow$ & $\delta_1\uparrow$ & AbsRel$\downarrow$ & $\delta_1\uparrow$ & AbsRel$\downarrow$ & $\delta_1\uparrow$ & AbsRel$\downarrow$ & $\delta_1\uparrow$ \\
    \midrule
DA~\cite{da}&11.1&83.5&5.4 &96.9&12.3&82.6&4.1 &98.3 &25.2&66.5&15.0&80.1&9.8 &89.1&12.5&81.9\\
DAV2~\cite{da2}&\underline{5.2}&\underline{97.2}&3.5 &\underline{99.4}&\underline{4.1}&\underline{99.5}&3.1 &99.5 &\textbf{23.0}  &\underline{68.1}&15.1& 79.8           & 10.3          & 89.0           &13.3&84.0\\
Metric3D~\cite{metric3d}&13.7&82.9&6.2&95.3&13.4&84.1&8.2&92.8 &71.2&22.5&---&--- &--- &--- &---& ---\\
Metric3D2~\cite{metric3d2}&6.1 &96.3&\textbf{2.9}&99.0&11.5&90.3&\underline{2.7}&99.3 &66.3&24.4& ---        & ---&--- &--- & ---           &---\\
Marigold~\cite{marigold}& 7.3  & 94.2 & 4.3  & 98.4 & 5.8  & 98.7  & 3.8  & 99.3 & 37.7  & 36.8 & 13.6  & 84.9 & 11.1  & 89.5  & 12.6  & 86.8\\
DepthFM~\cite{depthfm}& 8.3   & 92.6 & 4.8  & 98.3 & 10.5 & 88.9 & 6.1  & 95.3 & 54.2   & 26.9  & 17.0  & 76.4 & 13.3  & 83.2  & 14.4  & 81.9\\
Lotus~\cite{lotus}& 6.8  & 94.0 & 3.9  & 98.9 & 4.6  & 99.2 & 3.7  & \underline{99.6} & 47.1  & 30.3 & 13.5  & 85.4 & 10.6  & \underline{91.1} & \underline{11.3} & \textbf{89.6}\\
Depth4ToM~\cite{depth4tom}&6.0&96.6&5.0&97.9&5.0&97.4&4.0& 98.9 &---&--- &---&--- &---&--- &---& ---\\
RobustDepth~\cite{robustdepth}&---& ---   & ---   & ---   & ---   & ---& ---   & --- & 27.8  & 62.2 & 16.8   & 77.5  & 10.5   & 88.2  & 16.7   & 75.5 \\
VPD~\cite{vpd} & 5.5  & 95.9 & 4.2  & 98.6 & 5.2  & 98.4 & 3.6  & 99.2 & 31.5  & 63.1 & \underline{11.5}  & \underline{86.6} & \underline{9.2}  & 89.6 & 11.6  & 85.2\\
WorDepth~\cite{wordepth}    & 8.1  & 91.4 & 5.9  & 96.8 & 6.0  & 97.9 & 4.6  & 99.1 & 26.7  & 67.8 & 13.0  & 83.2 & 9.4   & 89.1 & 11.5  & 84.7\\
Ours    & \textbf{3.9} & \textbf{98.7} & \underline{3.3} & \textbf{99.5} & \textbf{3.8} & \textbf{99.6} & \textbf{2.1} & \textbf{99.9} & \underline{24.9} & \textbf{68.3} & \textbf{10.1} & \textbf{88.3} & \textbf{7.1} & \textbf{93.3} & \textbf{10.2} & \underline{88.2}\\
    \bottomrule
    \end{tabular}
\end{table*}

\begin{figure*}[!t]
\vspace{-5pt}
  \centering
  \includegraphics[width=\linewidth]{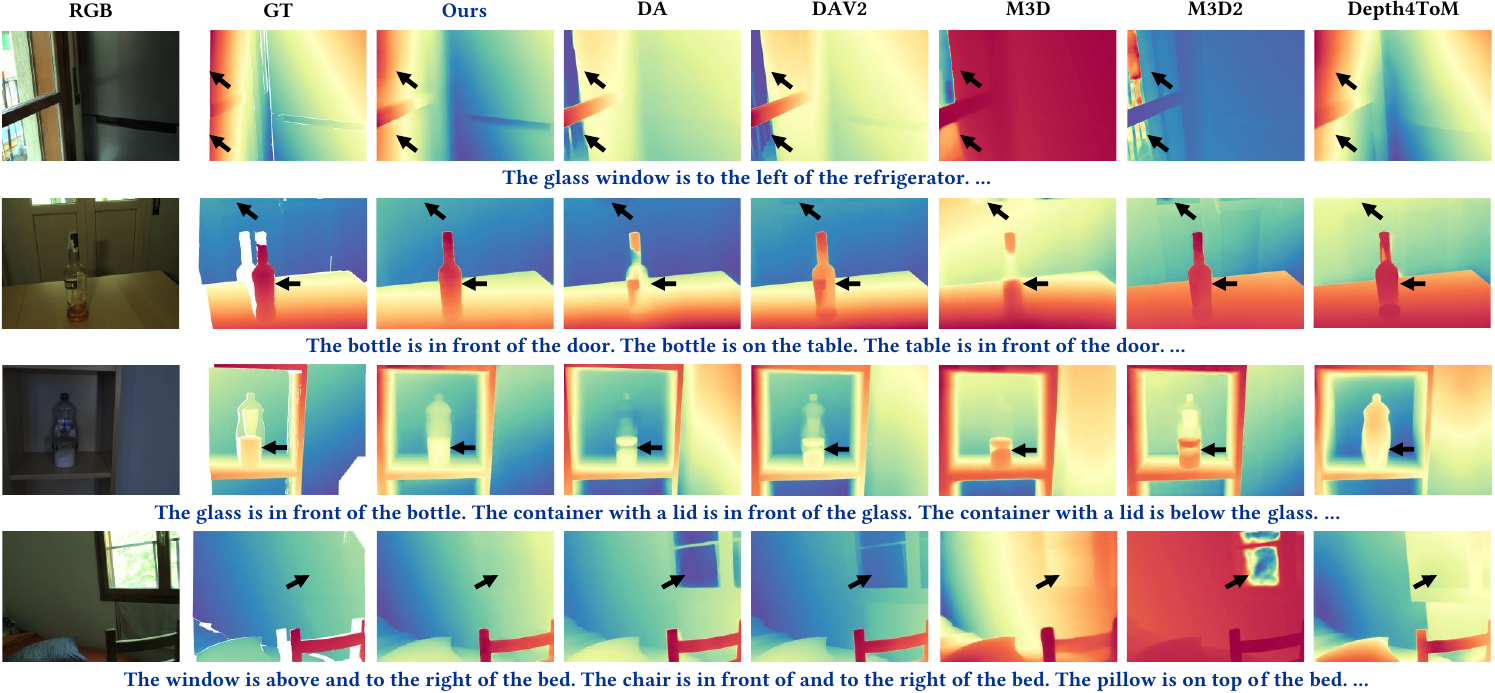}
  \vspace{-20pt}
  \caption{
Zero-shot comparisons in non-Lambertian surfaces.
``M3D'' denotes Metric3D.
CapDepth effectively leverages language guidance, achieving more robust MDE results in the arrow-highlighted regions.
}
  \vspace{-10pt}
  \label{fig:vis1}
\end{figure*}

\section{Experiments}
\label{sec:exp}

In this section, we first introduce the datasets and evaluation protocols in Sec.~\ref{sec:dataset}.
Implementation details are provided in Sec.~\ref{sec:details}.
Then we provide comparisons with state-of-the-art methods in Sec.~\ref{sec:compare}.
We also perform comprehensive ablation studies in Sec.~\ref{sec:ab}.

\subsection{Datasets and Evaluation Protocols}
\label{sec:dataset}
Following prior works \cite{marigold,depthfm,lotus}, we conduct our training on two datasets: Hypersim \cite{hypersim} and Virtual KITTI 2 \cite{vkitti2}.
We adopt data splits from Marigold \cite{marigold}, comprising 74K samples from Hypersim, along with 42K samples from Virtual KITTI 2.
We perform zero-shot test on four benchmarks: Booster \cite{booster}, ClearGrasp \cite{cleargrasp}, nuScenes \cite{nuscenes}, and DrivingStereo \cite{drivingstereo}.
Specifically, we evaluate on the official training split of Booster with 228 samples, the official real-test split of ClearGrasp with 286 samples, and the night-rain split of nuScenes with 120 samples.
DrivingStereo is officially partitioned into three subsets representing different adverse weather conditions—foggy, cloudy, and rainy—with each subset containing 500 samples.

Following prior works \cite{da2,marigold}, we employ two primary metrics: (1) the absolute relative error (AbsRel), defined as $\frac{1}{m}\sum_{i=1}^m\left|\frac{d_i-\hat{d}_i}{\hat{d_i}}\right|$,
and (2) the $\delta_1$ accuracy, defined as $\frac{1}{m}\sum_{i=1}^m\mathbb{I}\left(\max\left(\frac{d_i}{\hat{d}_i},\frac{\hat{d}_i}{d_{i}}\right)<1.25\right)$, where $\mathbb{I}(\cdot)$ represents the indicator function, $m$ denotes the number of the valid pixels, $\hat{d}$ represents the ground truth depth value of the pixel, and $d$ denotes the predicted depth value of the pixel.
We align $d$ and $\hat{d}$ to the same scale and shift following the prevalent method \cite{midas} before computing AbsRel and $\delta_1$.
For Booster \cite{booster} and ClearGrasp \cite{cleargrasp}, leveraging the officially provided segmentation masks, we additionally report AbsRel and $\delta_1$ on transparent or mirror (ToM) surfaces to enable comprehensive comparisons in non-Lambertian surfaces.
All metrics are reported as percentages.

\begin{figure}[!t]
  \centering
  \includegraphics[width=\linewidth]{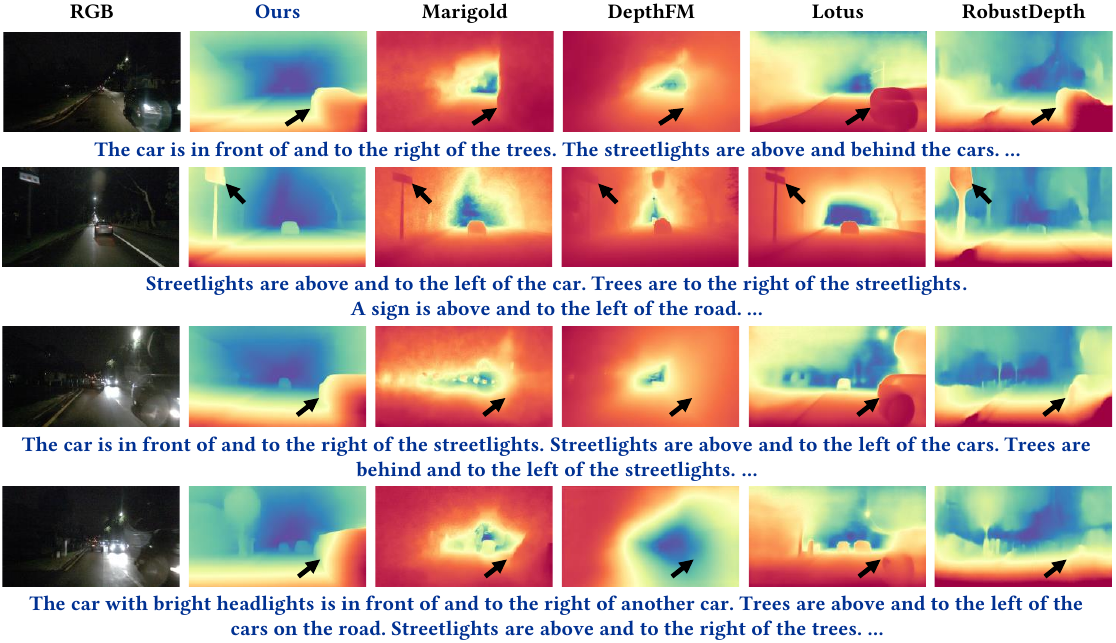}
  \vspace{-20pt}
  \caption{
Zero-shot comparisons in adverse weather conditions.
Our proposed CapDepth predicts more robust MDE results in the arrow-highlighted regions.
}
  \label{fig:vis3}
\end{figure}

\begin{figure}[!t]
\vspace{-5pt}
  \centering
  \includegraphics[width=\linewidth]{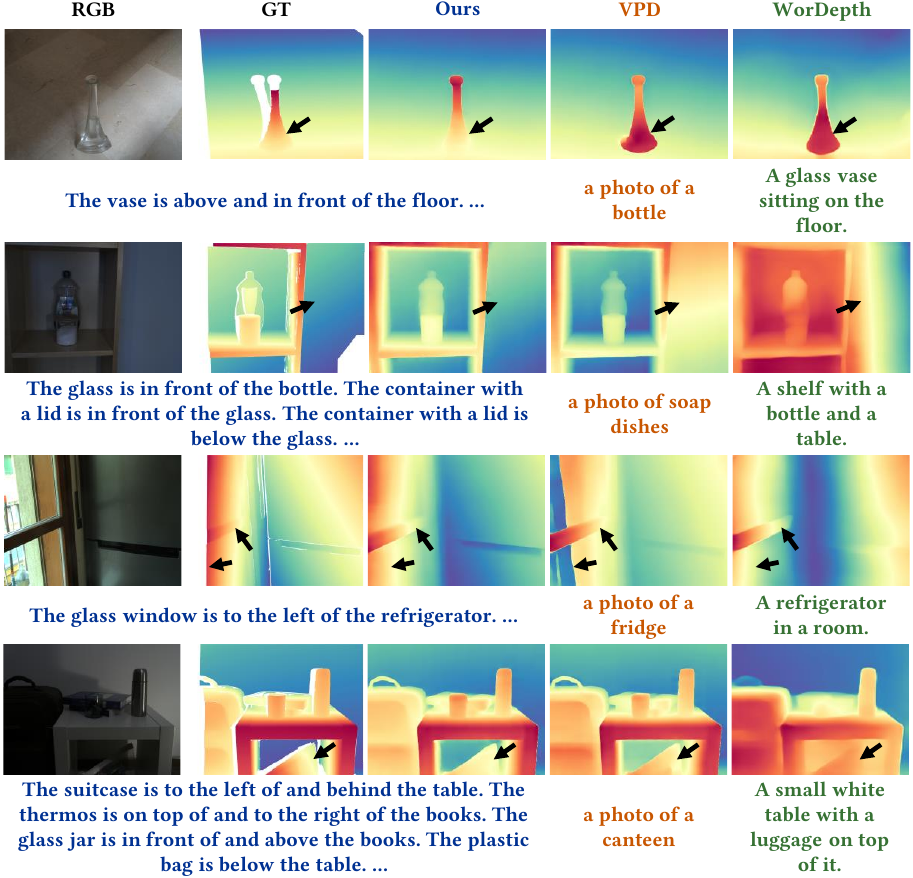}
  \vspace{-20pt}
  \caption{
Zero-shot comparisons in non-Lambertian surfaces.
We provide more qualitative comparisons with various methods in the supplementary material.
}
  \label{fig:vis2}
  \vspace{-5pt}
\end{figure}

\begin{figure}[!t]
  \centering
  \includegraphics[width=\linewidth]{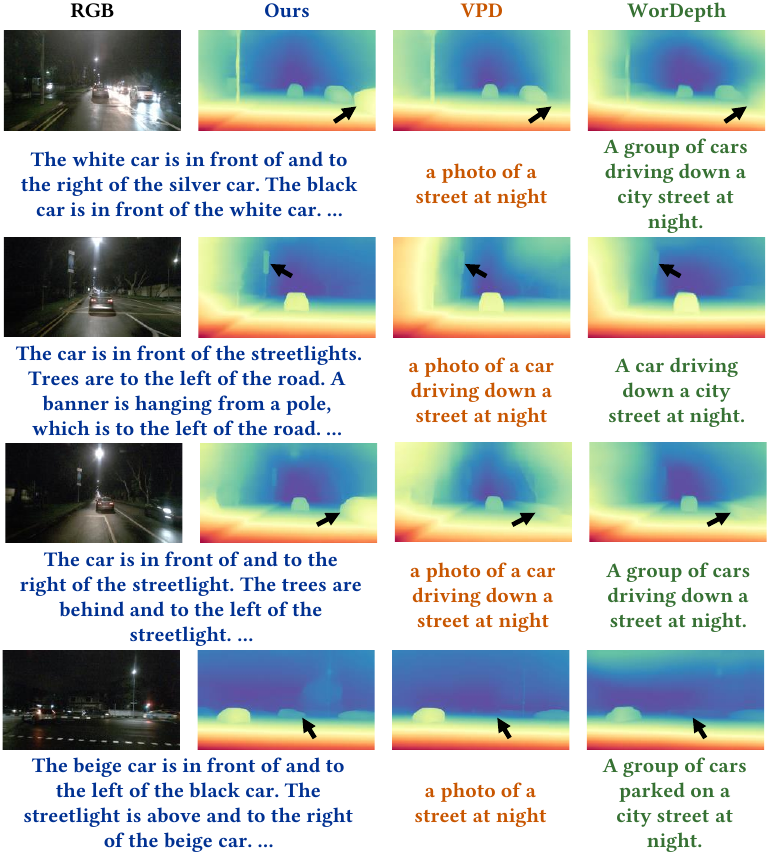}
  \vspace{-20pt}
  \caption{
Zero-shot comparisons in adverse weather conditions.
CapDepth better alleviates visual ambiguities and achieves more robust MDE in challenging scenarios by effectively leveraging guidance from detailed long caption input.
}
  \vspace{-15pt}
  \label{fig:vis4}
\end{figure}

\subsection{Implementation Details}
\label{sec:details}
We train our model for 5 epochs with a batch size of 32, distributed across 4 NVIDIA A6000 GPUs.
The number of the progressive masked querying blocks $L$ is set to 6.
$N$ in the \decoder~is set to 3.
We employ the AdamW optimizer with an initial learning rate of $3\times10^{-5}$ and a weight decay of $1\times10^{-2}$.
The regularization loss weight is set to $1\times10^{-3}$.
We apply the same data augmentation strategies as in the previous work \cite{vpd}.
More details are provided in the supplementary material.

\subsection{Comparisons to State-of-the-Art Methods}
\label{sec:compare}

\subsubsection{Quantitative Results}
As demonstrated in Tab.~\ref{tab:main}, CapDepth achieves state-of-the-art performance, surpassing prevalent general MDE methods (\eg, Depth Anything \cite{da} and Metric3D \cite{metric3d}), prior robust MDE works (\eg, Depth4ToM \cite{depth4tom} and RobustDepth \cite{robustdepth}), and existing language-integrated MDE models (\eg, VPD \cite{vpd} and WorDepth \cite{wordepth}).
To be specific, CapDepth reduces AbsRel by 25.0\% compared to Depth Anything V2 (DAV2) \cite{da2} and by 35.0\% compared to Depth4ToM \cite{depth4tom} on transparent or mirror (ToM) surfaces on Booster \cite{booster}.
Moreover, on nuScenes \cite{nuscenes}, our method attains a 6.1\% absolute improvement in $\delta_1$ accuracy over RobustDepth \cite{robustdepth}, a method explicitly designed for adverse weather conditions.
On the other hand, despite integrating simple short text input, VPD \cite{vpd} falls behind CapDepth by a 5.2\% absolute margin in $\delta_1$ accuracy when evaluated on rainy night scenes in nuScenes \cite{nuscenes}.
Furthermore, CapDepth attains a substantial 24.5\% reduction in AbsRel compared to WorDepth \cite{wordepth} on foggy day scenarios in DrivingStereo \cite{drivingstereo}.
Notably, on Booster \cite{booster}, CapDepth demonstrates a remarkable 51.9\% reduction in AbsRel on ToM surfaces relative to WorDepth \cite{wordepth}.
These quantitative improvements validate the effectiveness of incorporating detailed long captions, extracting fine-grained depth-relevant text features, and enhancing depth decoding with language guidance, which enables more robust MDE across both non-Lambertian surfaces and adverse weather conditions.

\subsubsection{Qualitative Results}
As illustrated in Figs.~\ref{fig:vis1}~to~\ref{fig:vis4}, through our effective design of the \template, the \encoder, and the \decoder, CapDepth achieves more robust MDE results compared to previous methods \cite{da,da2,metric3d,metric3d2,marigold,lotus,depthfm,depth4tom,robustdepth,vpd,wordepth} both on non-Lambertian surfaces and under adverse weather conditions.
Specifically, despite employing scenario-specific data augmentations, Depth4ToM \cite{depth4tom} and RobustDepth \cite{robustdepth} exhibit limited robustness improvements on non-Lambertian surfaces (Fig.~\ref{fig:vis1}) and under adverse weather conditions (Fig.~\ref{fig:vis3}), respectively.
However, CapDepth can capture a more accurate depth for transparent windows and bottles (Fig.~\ref{fig:vis1}) and the car in the nighttime scene (Fig.~\ref{fig:vis3}) through language guidance.
On the other hand, despite incorporating language, VPD \cite{vpd} and WorDepth \cite{wordepth} exhibit suboptimal MDE performance in these challenging scenarios (Figs.~\ref{fig:vis2}~to~\ref{fig:vis4}).
This stems from their reliance on simple short text input, coarse global text feature extraction, and limited text guidance during depth decoding.
In contrast, CapDepth leverages rich and precise spatial information from language modality, enabling robust MDE with more accurate geometric structure across both non-Lambertian surfaces (Fig.~\ref{fig:vis2}) and adverse weather conditions (Fig.~\ref{fig:vis4}), where RGB information alone is insufficient to alleviate visual ambiguities caused by these challenging scenarios.

\begin{table}[!b]
\vspace{-5pt}
\caption{
Efficacy of CapDepth.
We evaluate the proposed \encoder~and~\decoder~in non-Lambertian surfaces (ToM surfaces on Booster \cite{booster}) and adverse weather conditions (rainy night on nuScenes \cite{nuscenes}) by removing $L$ progressive masked querying blocks or $N$ SAdaLN.
}
\vspace{-5pt}
\label{tab:ab}
\setlength{\tabcolsep}{2pt}
\begin{tabular}{cccccc}
    \toprule
Dynamic Caption & Text-Adaptive & \multicolumn{2}{c}{ToM} & \multicolumn{2}{c}{night-rain} \\
Encoder       & Decoder & AbsRel$\downarrow$ & $\delta_1\uparrow$ & AbsRel$\downarrow$ & $\delta_1\uparrow$ \\
    \midrule
\ding{55} & \ding{51}    & 8.1 & 89.1 & 44.7 & 47.6  \\
\ding{51} & \ding{55}    & 5.0 & 97.4 & 38.4 & 59.2  \\
\ding{51} & \ding{51}    & \textbf{3.9} & \textbf{98.7} & \textbf{24.9} & \textbf{68.3}  \\
    \bottomrule
\end{tabular}
\end{table}

\begin{table}[!b]
\caption{
Analysis of the impact of various text inputs on different language-integrated MDE models in non-Lambertian surfaces (ToM surfaces on Booster \cite{booster}) and adverse weather conditions (rainy night on nuScenes \cite{nuscenes}).
}
\vspace{-5pt}
\label{tab:ab_text}
\setlength{\tabcolsep}{3pt}
\begin{tabular}{cccccc}
    \toprule
\multirow{2}{*}{Method} & \multirow{2}{*}{Text Input} & \multicolumn{2}{c}{ToM} & \multicolumn{2}{c}{night-rain} \\
& & AbsRel$\downarrow$ & $\delta_1\uparrow$ & AbsRel$\downarrow$ & $\delta_1\uparrow$ \\
    \midrule
\multirow{3}{*}{VPD \cite{vpd}}
&VPD \cite{vpd}             & 5.5  & 95.9 & 31.5 & 63.1 \\
& WorDepth \cite{wordepth}  & 11.1 & 91.7 & 51.8 & 35.2 \\
& CapDepth                  & 11.1 & 91.6 & 51.8 & 35.1 \\
    \midrule
\multirow{3}{*}{WorDepth \cite{wordepth}}
& VPD \cite{vpd}            & 8.0 & 90.6 & 26.9 & 66.3  \\
& WorDepth \cite{wordepth}  & 8.1 & 91.4 & 26.7 & 67.8  \\
& CapDepth                  & 8.1 & 90.5 & 27.0 & 66.3  \\
    \midrule
\multirow{3}{*}{CapDepth}
& VPD \cite{vpd}            & 4.3 & 98.1 & 25.7 & 67.9  \\
& WorDepth \cite{wordepth}  & 4.4 & 97.5 & 25.7 & 67.9  \\
& CapDepth & \textbf{3.9} & \textbf{98.7} & \textbf{24.9} & \textbf{68.3}  \\
    \bottomrule
\end{tabular}
\end{table}

\subsection{Ablation Studies}
\label{sec:ab}

\subsubsection{Efficacy of CapDepth}
\label{sec:ab_main}
In Tab.~\ref{tab:ab}, we analyze the effectiveness of our designed \encoder~and \decoder.
The quantitative results reveal that directly utilizing text features from CLIP results in a 9.6\% absolute degradation in $\delta_1$ on non-Lambertian surfaces.
This indicates that original CLIP features, which are optimized for the image-text retrieval task \cite{clip}, lack fine-grained depth-relevant information necessary for effective MDE guidance.
Furthermore, when removing the SAdaLN in the \decoder, we observe a 54.2\% increase in AbsRel under adverse weather conditions.
This validates the existence of the gap between U-Net noise prediction features and the precise geometric reasoning required for MDE \cite{evp} as discussed in Sec.~\ref{sec:overview}.

\subsubsection{Discussion of Text Input Template}
\label{sec:template}
In Tab.~\ref{tab:ab_text}, we analyze the impact of various text inputs on different language-integrated MDE methods.
The quantitative results demonstrate two points:
(1) Coarse global text feature extraction of previous works \cite{vpd,wordepth} discards information from detailed long caption input, leading to suboptimal MDE results;
(2) Since text inputs of VPD \cite{vpd}, WorDepth \cite{wordepth}, and CapDepth can be considered to contain varying quality of spatial information, results indicate that our CapDepth can maintain robust performance when provided with lower-quality text inputs (\eg, those from VPD \cite{vpd} and WorDepth \cite{wordepth}).

\subsubsection{Analysis of Atom Sentence}
\begin{figure}[!t]
\vspace{-10pt}
  \centering
  \includegraphics[width=\linewidth]{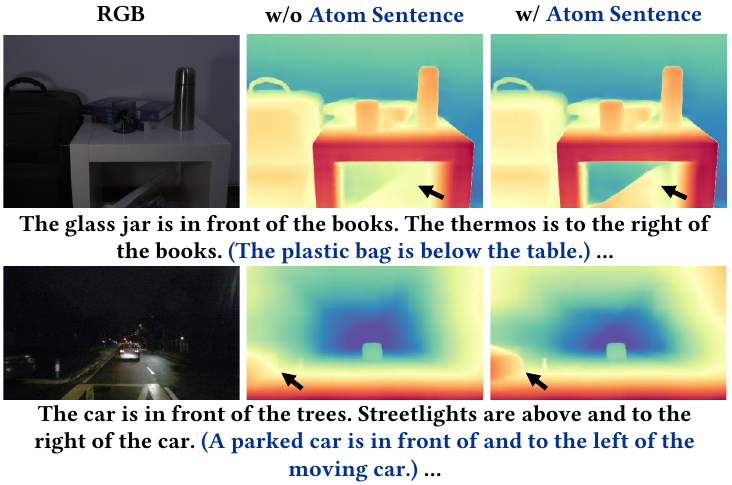}
\vspace{-20pt}
  \caption{
Visual analysis of atom sentences.
We conduct this on individual atom sentences by deleting the certain one.
}
  \label{fig:ab}
\vspace{-10pt}
\end{figure}

In Tab.~\ref{tab:ab_sentence}, we analyze the impact of the atom sentence quantity on guiding MDE at the dataset level.
The quantitative results reveal a positive correlation between the number of atom sentences and overall model performance. 
Furthermore, we qualitatively investigate the guiding role of the individual atom sentence in Fig.~\ref{fig:ab}.
By incorporating the specific atom sentence that explicitly describes the spatial locations of the ambiguous objects—such as the transparent plastic bag (Fig.~\ref{fig:ab}~(up)) or the car in the nighttime scene (Fig.~\ref{fig:ab}~(down))—CapDepth demonstrates an enhanced capability to accurately perceive their depth across both non-Lambertian surfaces and adverse weather conditions.

\begin{table}[!t]
\caption{
Analysis of different numbers of atom sentences in non-Lambertian surfaces (ToM surfaces on Booster \cite{booster}) and adverse weather conditions (rainy night on nuScenes \cite{nuscenes}).
}
\vspace{-5pt}
\label{tab:ab_sentence}
\setlength{\tabcolsep}{7pt}
\begin{tabular}{ccccc}
\toprule
The Number of & \multicolumn{2}{c}{ToM} & \multicolumn{2}{c}{night-rain} \\
Atom Sentences & AbsRel$\downarrow$ & $\delta_1\uparrow$ & AbsRel$\downarrow$ & $\delta_1\uparrow$ \\
\midrule
0       & 4.32 & 98.22 & 28.21 & 65.07  \\
1       & 4.07 & 98.61 & 25.65 & 67.79  \\
2       & 3.97 & 98.66 & 25.14 & 68.25  \\
$\geq$3 & \textbf{3.96} & \textbf{98.70} & \textbf{24.99} & \textbf{68.30}  \\
\bottomrule
\end{tabular}
\end{table}

\section{Conclusions}
We propose CapDepth, a novel framework leveraging detailed long captions for robust monocular depth estimation (MDE) across both non-Lambertian surfaces and adverse weather conditions via:
(1) a \template~to provide rich spatial information,
(2) a \encoder~for fine-grained depth-relevant text feature,
(3) a \decoder~that guides enhanced depth decoding.
CapDepth validates that detailed long captions can guide more robust visual perception.
However, CapDepth only explores English texts.
Future work can investigate generalization across diverse languages and conduct systematic cross-lingual comparisons to assess language-specific performance variations.

\bibliographystyle{ACM-Reference-Format}
\balance
\bibliography{main}

\newpage
\appendix
\renewcommand{\thesection}{\Alph{section}}
\renewcommand{\thefigure}{\Alph{figure}}
\renewcommand{\thetable}{\Alph{table}}
\setcounter{section}{0}
\setcounter{figure}{0}
\setcounter{table}{0}

\section{More Experimental Results}

\subsection{Detailed Long Caption Input Examples}
\begin{figure}[b]
  \centering
  \includegraphics[width=\linewidth]{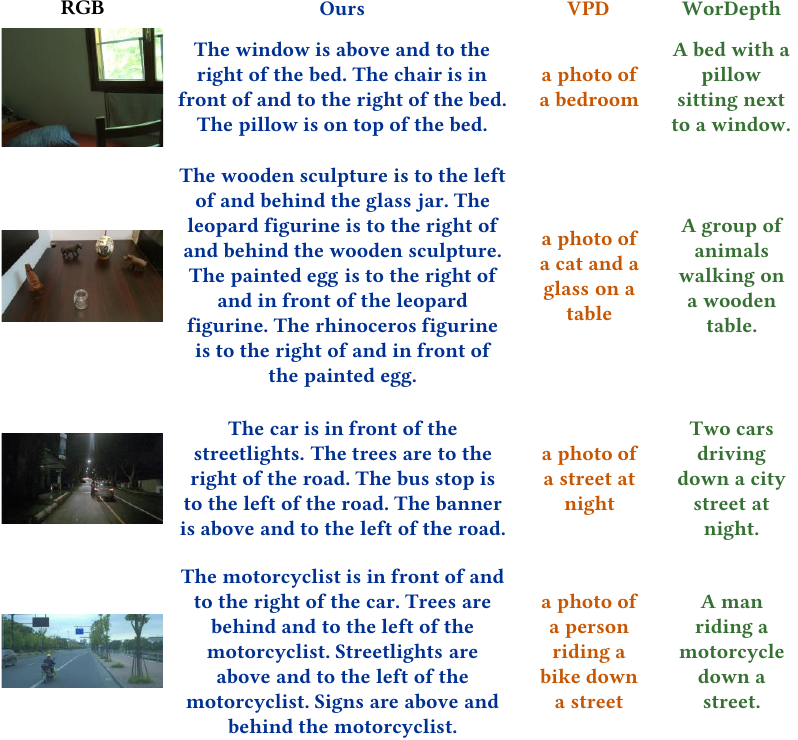}
  \caption{
Detailed long caption input examples and comparisons with prior language-integrated MDE methods \cite{vpd,wordepth}.
}
  \label{fig:dlc_example}
\end{figure}

We provide a comprehensive illustration of our detailed long caption input samples and comparisons with other language-integrated MDE methods \cite{vpd,wordepth} in Fig.~\ref{fig:dlc_example}.
The detailed long captions offer richer and more precise guidance for robust MDE in challenging scenarios compared to the simple short texts used by the previous language-integrated MDE methods \cite{vpd,wordepth}.

\subsection{More Qualitative Comparisons}
\begin{figure}[t]
  \centering
  \includegraphics[width=\linewidth]{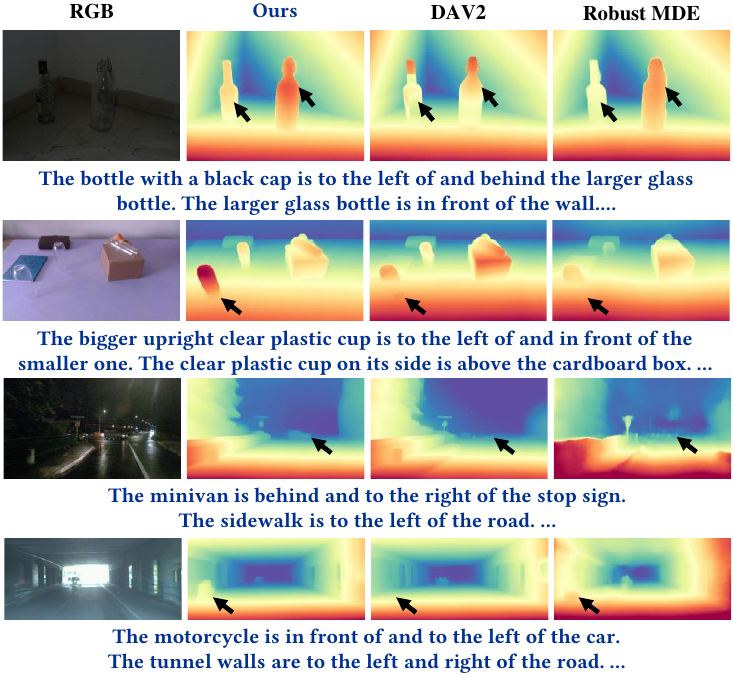}
  \caption{
Additional qualitative comparisons.
CapDepth (Ours) outperforms the general MDE method (Depth Anything V2 \cite{da2}, abbreviated as DAV2) and the robust MDE methods (Depth4Tom \cite{depth4tom} for the first two rows and RobustDepth \cite{robustdepth} for the last two rows) in multiple similar transparent bottles (even under low light conditions), low-visibility nighttime vehicles, and objects with motion-induced blur (fast-moving motorcycles).
}
  \label{fig:supp_vis1}
\end{figure}

\begin{figure}[t]
  \centering
  \includegraphics[width=\linewidth]{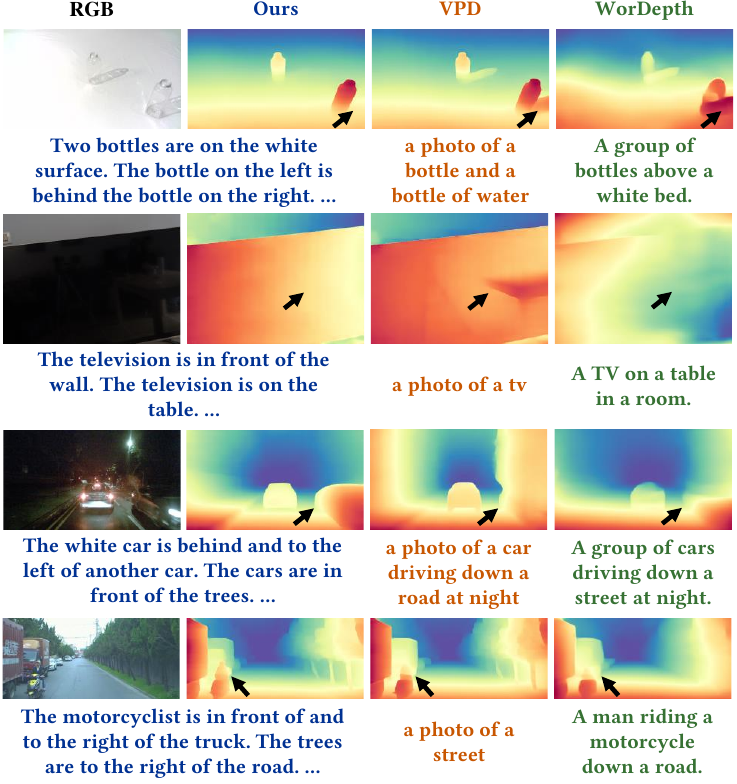}
  \caption{
Additional qualitative comparisons.
CapDepth (Ours) outperforms previous language-integrated MDE methods (VPD \cite{vpd} and WorDepth \cite{wordepth}) in multiple similar transparent bottles, reflective televisions, black cars in nighttime scenes, and moving motorcyclists.
}
  \label{fig:supp_vis4}
\end{figure}

\begin{figure*}[p]
  \centering
  \begin{minipage}[c][0.92\textheight][c]{\linewidth}
    \centering
    \includegraphics[width=\linewidth]{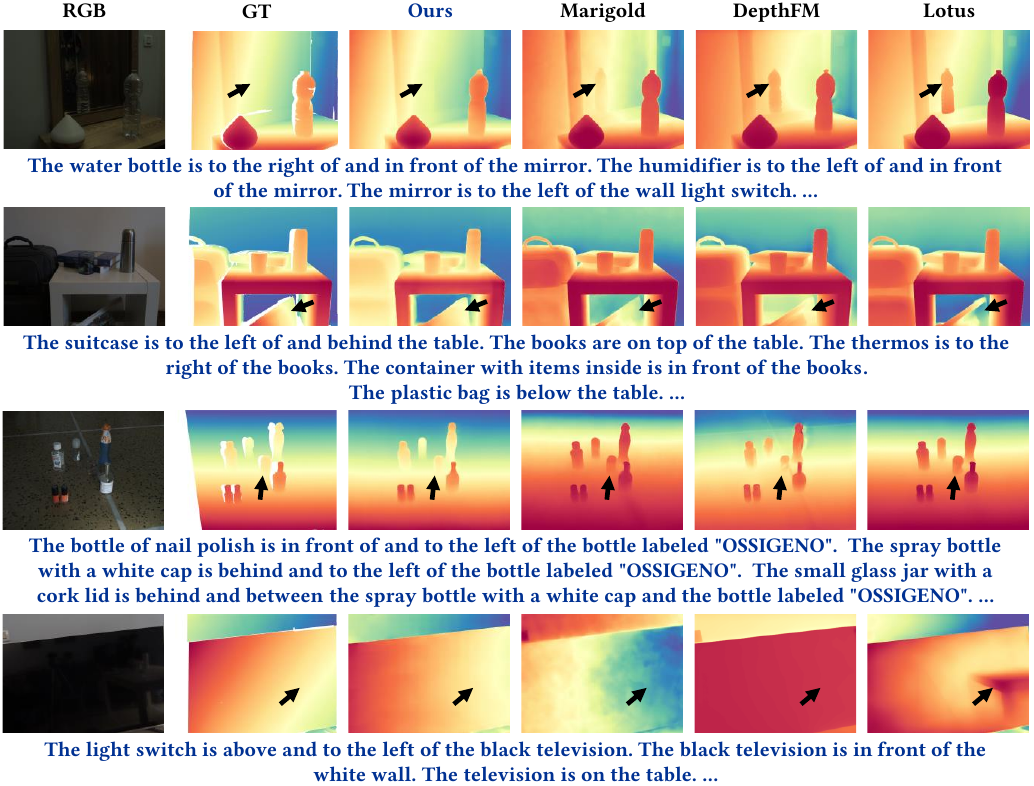}
    \caption{
Additional qualitative comparisons.
CapDepth (Ours) outperforms the general MDE methods (Marigold \cite{marigold}, DepthFM \cite{depthfm}, and Lotus \cite{lotus}) in reflective mirrors, transparent plastic bags, transparent bottles, and reflective television screens.
    }
    \label{fig:supp_vis2}
  \end{minipage}
\end{figure*}

\begin{figure*}[p]
  \centering
  \begin{minipage}[c][0.92\textheight][c]{\linewidth}
    \centering
    \includegraphics[width=\linewidth]{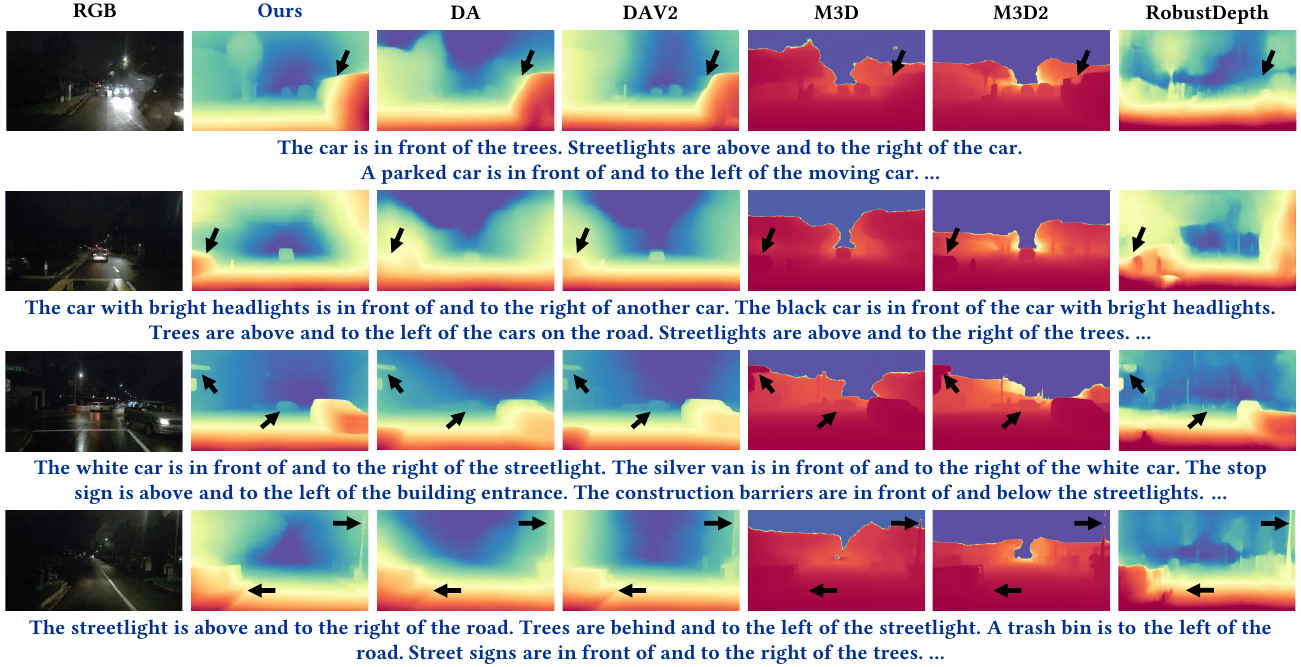}
    \caption{
Additional qualitative comparisons.
CapDepth (Ours) outperforms general MDE methods (Depth Anything \cite{da}, abbreviated as DA, Depth Anything V2 \cite{da2}, abbreviated as DAV2, Metric3D \cite{metric3d}, abbreviated as M3D, Metric3D V2 \cite{metric3d2}, abbreviated as M3D2) and robust MDE method (RobustDepth \cite{robustdepth}) in nighttime scenes with moving cars and road sign poles.
    }
    \label{fig:supp_vis3}
  \end{minipage}
\end{figure*}

\begin{figure*}[p]
  \centering
  \begin{minipage}[c][0.92\textheight][c]{\linewidth}
    \centering
    \includegraphics[width=\linewidth]{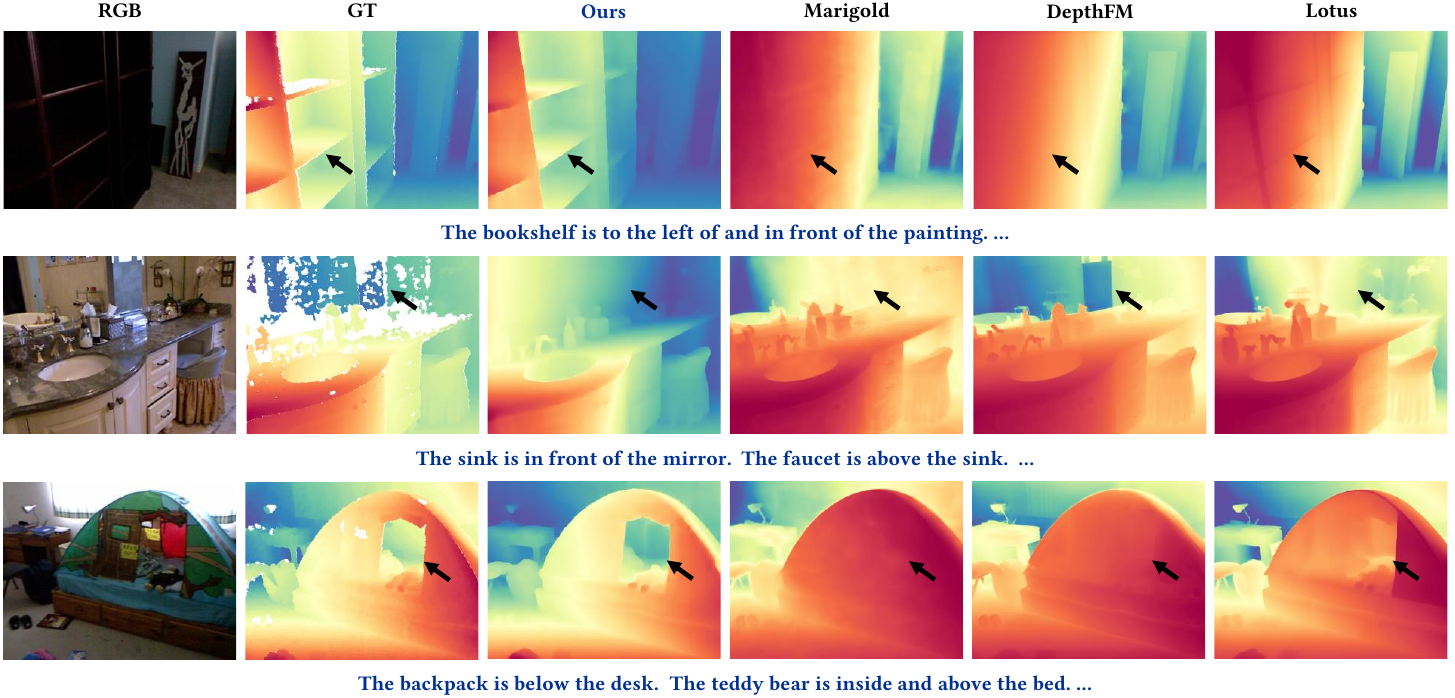}
    \caption{
Additional qualitative comparisons.
CapDepth (Ours) outperforms general MDE methods (Marigold \cite{marigold}, DepthFM \cite{depthfm}, and Lotus \cite{lotus}) in weakly-textured regions (first row), reflective mirrors (second row), and complex-textured regions (third row).
    }
    \label{fig:supp_vis6}
  \end{minipage}
\end{figure*}

\begin{figure*}[p]
  \centering
  \begin{minipage}[c][0.92\textheight][c]{\linewidth}
    \centering
    \includegraphics[width=\linewidth]{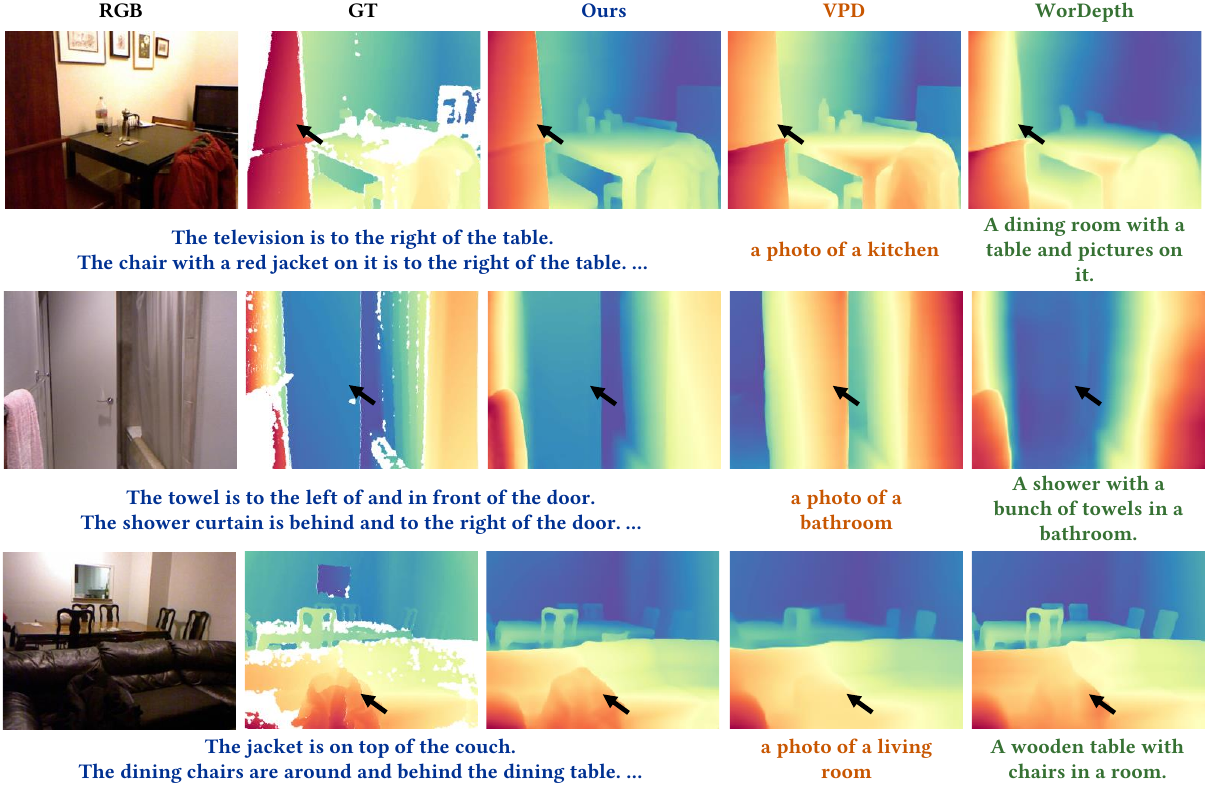}
    \caption{
Additional qualitative comparisons.
CapDepth (Ours) outperforms previous language-integrated MDE methods (VPD \cite{vpd} and WorDepth \cite{wordepth}) in weakly-textured or complex-textured areas, as highlighted by the arrows.
    }
    \label{fig:supp_vis7}
  \end{minipage}
\end{figure*}

We provide additional qualitative comparisons to further demonstrate that our CapDepth outperforms previous state-of-the-art baselines, including general MDE methods (Depth Anything \cite{da}, abbreviated as DA, Depth Anything V2 \cite{da2}, abbreviated as DAV2, Metric3D \cite{metric3d}, abbreviated as M3D, Metric3D V2 \cite{metric3d2}, abbreviated as M3D2, Marigold \cite{marigold}, DepthFM \cite{depthfm}, and Lotus \cite{lotus}), robust MDE approaches (Depth4ToM \cite{depth4tom} and RobustDepth \cite{robustdepth}), and language-integrated MDE methods (VPD \cite{vpd} and WorDepth \cite{wordepth}) in Figs. \ref{fig:supp_vis1} to \ref{fig:supp_vis7}.
Results show that our CapDepth effectively leverages language guidance to improve the robustness of MDE across non-Lambertian surfaces and adverse weather conditions, thereby alleviating the visual ambiguities caused by these challenging scenarios.

\subsection{More Quantitative Comparisons}
We provide additional quantitative comparisons to further demonstrate that our CapDepth outperforms previous state-of-the-art works, including general MDE methods (Depth Anything \cite{da}, abbreviated as DA, Depth Anything V2 \cite{da2}, abbreviated as DAV2, Metric3D \cite{metric3d}, Metric3D2 \cite{metric3d2}, Marigold \cite{marigold}, DepthFM \cite{depthfm}, and Lotus \cite{lotus}), robust MDE methods (Depth4ToM \cite{depth4tom} and RobustDepth \cite{robustdepth}), and language-integrated MDE methods (VPD \cite{vpd} and WorDepth \cite{wordepth}) in Tab.~\ref{tab:supp_main}.
Specifically, we report additional evaluation metrics beyond those presented in Table 2 of the main paper, including the squared relative error (SqRel), defined as
$\frac{1}{m}\sum_{i=1}^{m}\frac{(d_i-\hat{d}_i)^2}{\hat{d}_i}$,
and the root mean squared error (RMSE), defined as
$\sqrt{\frac{1}{m}\sum_{i=1}^{m}(d_i-\hat{d}_i)^2}$, where $m$ denotes the number of the valid pixels, $\hat{d}$ represents the ground truth depth value of the pixel, and $d$ denotes the predicted depth value of the pixel.
The results show that our CapDepth effectively leverages language guidance to improve the robustness of MDE across both non-Lambertian surfaces and adverse weather conditions, outperforming previous state-of-the-art baselines.

\section{More Experimental Details}
\label{sec:detail}

\subsection{Derivations for $\mathcal{L}_{reg}$}

$$
\begin{aligned}
\mathcal{L}_{reg} &= D_{KL}(\mathcal{N}(b, a^2I) || \mathcal{N}(0, I)) \\
&\underset{\text{Wise}}{\overset{\text{Channel}}{=}} \sum_{c} D_{KL}(\mathcal{N}(b_c, a_c^2) || \mathcal{N}(0, 1)) \\
&= \sum_{c} \int \mathcal{N}(x_c; b_c, a_c^2) \log \frac{\mathcal{N}(x_c; b_c, a_c^2)}{\mathcal{N}(x_c; 0, 1)} dx_c \\
&= \sum_{c} \mathbb{E}_{x_c \sim \mathcal{N}(b_c, a_c^2)} \Bigg[ \log \left( \frac{1}{\sqrt{2\pi a_c^2}} e^{-\frac{(x_c-b_c)^2}{2a_c^2}} \right)
\\ &\qquad\qquad\qquad\qquad - \log \left( \frac{1}{\sqrt{2\pi}} e^{-\frac{x_c^2}{2}} \right) \Bigg] \\
&= \sum_{c} \mathbb{E}_{x_c \sim \mathcal{N}(b_c, a_c^2)} \left[ -\frac{1}{2}\log(a_c^2) - \frac{(x_c-b_c)^2}{2a_c^2} + \frac{x_c^2}{2} \right] \\
&= \sum_{c} \left( -\frac{1}{2}\log(a_c^2) - \frac{1}{2a_c^2}\mathbb{E}_{x_c}[(x_c-b_c)^2] + \frac{1}{2}\mathbb{E}_{x_c}[x_c^2] \right) \\
&= \sum_{c} \left( -\log|a_c| - \frac{1}{2} + \frac{a_c^2 + b_c^2}{2}\right) \\
&\underset{\text{Notation}}{\overset{\text{Vectorized}}{=}} -\log |a|+\frac{a^2+b^2}{2} -\frac{1}{2}
\end{aligned}
$$

\subsection{More Implementation Details}

\begin{table*}[!t]
\caption{
Additional quantitative comparisons.
DA denotes Depth Anything.
``---'' means that Metric3D models are trained on DrivingStereo (violating the zero-shot setting), and Depth4ToM and RobustDepth are designed for non-Lambertian surfaces and adverse weather conditions, respectively, making them inapplicable to the other challenging scenario.
}
\label{tab:supp_main}
    \setlength{\tabcolsep}{0.8pt}
    \begin{tabular}{ccccccccccccccccccccccccc}
    \toprule
\multirow{3}{*}{Method} & \multicolumn{4}{c}{Booster \cite{booster}} & \multicolumn{4}{c}{ClearGrasp \cite{cleargrasp}} & \multicolumn{2}{c}{nuScenes \cite{nuscenes}} & \multicolumn{6}{c}{DrivingStereo \cite{drivingstereo}}\\
& \multicolumn{2}{c}{ToM} & \multicolumn{2}{c}{All} & \multicolumn{2}{c}{ToM} & \multicolumn{2}{c}{All} & \multicolumn{2}{c}{night-rain} & \multicolumn{2}{c}{cloudy} & \multicolumn{2}{c}{foggy} & \multicolumn{2}{c}{rainy} \\
& SqRel$\downarrow$ & RMSE$\downarrow$ & SqRel$\downarrow$ & RMSE$\downarrow$ & SqRel$\downarrow$ & RMSE$\downarrow$ & SqRel$\downarrow$ & RMSE$\downarrow$ & SqRel$\downarrow$ & RMSE$\downarrow$ & SqRel$\downarrow$ & RMSE$\downarrow$ & SqRel$\downarrow$ & RMSE$\downarrow$ & SqRel$\downarrow$ & RMSE$\downarrow$ \\
    \midrule
DA~\cite{da}&10.228&58.583&2.4967&28.092&0.0380&0.2345&0.0060&0.0891&0.0077&0.0229&0.0015&\underline{0.0117}&0.0008&\textbf{0.0071}&\underline{0.0013}&0.0128\\
DAV2~\cite{da2}&2.4116&26.885&\underline{1.0186} &\underline{18.069}&0.0153&0.1247&0.0032&0.0615 &\underline{0.0074}  &\underline{0.0223}&\underline{0.0014}&0.0118&\underline{0.0007}&\underline{0.0072}&\textbf{0.0012}&\underline{0.0127}\\
Metric3D~\cite{metric3d}&45.027&152.77&12.676&97.759&0.0188&0.1191&0.0098&0.0729 &7.5194&11.374&---&--- &--- &--- &---& ---\\
Metric3D2~\cite{metric3d2}&20.267 &90.711&5.8404&50.641&0.0140&0.1106&\underline{0.0017}&\underline{0.0369}&6.9529&11.051& ---        & ---&--- &--- & ---           &---\\
Marigold~\cite{marigold}& 13.957  & 81.070 & 5.3587  & 63.243& 0.0141  & \underline{0.1105}  & 0.0025  & 0.0379 & 3.0606  & 8.4763  & 0.9523  & 4.9440 & 0.7503  &  5.0468  & 0.9309  & 5.4528 \\
DepthFM~\cite{depthfm}& 27.180   & 105.71& 7.7388  & 74.410& 0.0196 & 0.1295 & 0.0072  & 0.0563 & 4.8770   & 9.5453  & 1.3685  & 5.8236 & 0.9222  & 5.5199  & 1.1419  & 5.8981\\
Lotus~\cite{lotus}& 10.476  & 70.301 & 4.3016  & 55.984& \underline{0.0137}  & 0.1133 & 0.0019  & 0.0372 & 3.0775  & 8.1037 & 1.0122  & 5.0504 & 0.7447  & 4.9438 & 0.8737 & 5.1859\\
Depth4ToM~\cite{depth4tom}&\underline{1.9269}&\underline{25.271}&1.9798&23.563&0.0199&0.1169&0.0059& 0.0834 &---&--- &---&--- &---&--- &---& ---\\
RobustDepth~\cite{robustdepth}&---& ---   & ---   & ---   & ---   & ---& ---   & --- & 0.0094  & 0.0266 & 0.0019   & 0.0132  & 0.0009   & 0.0078  & 0.0019   & 0.0138 \\
VPD~\cite{vpd} & 3.1208  & 29.151 & 1.5909  & 20.603 & 0.0248  & 0.1546 & 0.0047  & 0.0669 & 0.0127  & 0.0260 & 0.0017  & 0.0126 & 0.0008  & 0.0077 & {0.0014}  & {0.0130}\\
WorDepth~\cite{wordepth}    & 5.1743  & 41.526 & 2.8886  & 30.061& 0.0203  & 0.1548 & 0.0066  & 0.0926 & 0.0098  & 0.0231 & 0.0021  & 0.0135 & 0.0009   & 0.0078 & 0.0015  & {0.0130}\\
Ours    & \textbf{1.4907 }& \textbf{21.031} & \textbf{0.8943} & \textbf{16.207} & \textbf{0.0133} & \textbf{0.1096 }& \textbf{0.0016} &  \textbf{0.0365 }& \textbf{0.0072} & \textbf{0.0222} & \textbf{0.0013} & \textbf{0.0116} & \textbf{0.0006} & \textbf{0.0071} & \textbf{0.0012} & \textbf{0.0125} \\
    \bottomrule
    \end{tabular}
\end{table*}

Following previous methods \cite{vpd,wordepth}, during the training phase, we incorporate the image augmentation from CutDepth \cite{cutdepth}, random brightness and contrast augmentations within [-0.2, 0.2], and random gamma augmentation within [80, 120], all with a 50\% probability.
Image inputs are resized to $512\times512$ following priors \cite{vpd,wordepth}.
For the text inputs of CapDepth, we use the InternVL2.5 \cite{internvl} to generate detailed long captions with this prompt:

\textit{
Generate descriptions of spatial relationships between all visible objects in the image in the format:\\
\indent \{object A\} [spatial relationship phrases] \{object B\}. \\ 
\indent Ensure that:\\
\indent 1. Spatial relationship phrases only include `behind', `in front of', `above', `below', `to the left of', `to the right of'.\\
\indent 2. Nothing unrelated to spatial relationships is output. \\
\indent Example: \\
\indent `The dog is in front of and to the right of the mirror. Black cars are behind and below and to the right of the trees.'}

Although we constrain the output to specific spatial relationship phrases in the prompt, the model can still generate a variety of spatial relationship phrases, thereby enhancing CapDepth's generalization to diverse spatial relationship phrases.
For Table 1 in the main paper, we use the InternVL2.5 \cite{internvl} to analyze the linguistic statistics of each text input on the Hypersim \cite{hypersim} dataset.

\section{More Ablation Studies}

\begin{table}[!b]
\caption{
Analysis of the robustness of our CapDepth to input texts of varying quality in non-Lambertian surfaces (ToM surfaces on Booster \cite{booster}) and adverse weather conditions (rainy night on nuScenes \cite{nuscenes}).
``Degraded'' refers to applying perturbations to the input text (\eg, changing ``to the left of'' to ``to the right of'', ``in front of'' to ``behind'', etc.) with a 50\% probability.
``Vanilla'' refers to using the input text without constraining it to follow our designed \template~(\ie, using prompts such as ``Describe this image.'' for vision-language models).
These results demonstrate that CapDepth maintains robustness under input texts of varying quality, and further confirm the effectiveness of the proposed \template.
}
\label{tab:ab_vlm}
\setlength{\tabcolsep}{3pt}
\begin{tabular}{ccccc}
\toprule
\multirow{2}{*}{Vision-Language Model} & \multicolumn{2}{c}{ToM} & \multicolumn{2}{c}{night-rain} \\
& AbsRel$\downarrow$ & $\delta_1\uparrow$ & AbsRel$\downarrow$ & $\delta_1\uparrow$ \\
\midrule
BLIP2~\cite{blip2}       & 4.3 & 98.1 & 25.7 & 67.9  \\
ExpansionNet2~\cite{expansionnet}       & 4.4 & 97.5 & 25.7 & 67.9  \\
InternVL2.5-1B~\cite{internvl}& 4.1 & 98.5 & 25.3 & 68.1  \\
InternVL2.5-8B~\cite{internvl}       & 4.0 & 98.6 & 25.1 & 68.2  \\
InternVL2.5-38B~\cite{internvl} & \textbf{3.9} & 98.6 & 25.0 & 68.2  \\
InternVL2.5-78B~\cite{internvl} (Ours) & \textbf{3.9} & \textbf{98.7} & \textbf{24.9} & \textbf{68.3}  \\
InternVL2.5-78B~\cite{internvl} (Degraded) & 4.4 & 97.8 & 25.6 & 68.0  \\
InternVL2.5-78B~\cite{internvl} (Vanilla) & 4.9 & 97.2 & 26.1 & 67.6  \\
\bottomrule
\end{tabular}
\end{table}

\subsection{Varying Text Quality}
We conduct an additional ablation study to evaluate the robustness of CapDepth to varying input text quality.
Specifically, we use different vision-language models (VLMs) with varying capabilities to generate text inputs that meet the requirements of our designed \template.
In addition, we introduce degradations into the input text to analyze the robustness of CapDepth to descriptions containing incorrect spatial relationships.
We also examine the impact on model performance when the input text does not conform to the requirements of the designed \template.
As shown in Tab.~\ref{tab:ab_vlm}, stronger VLMs lead to better model performance in challenging scenarios due to the text inputs they produced that better adhere to the \template requirements.
When the spatial relationships described in the input text are incorrect, CapDepth still maintains a certain degree of robustness.
However, when the input text does not follow the \template, the conveyed spatial information becomes limited, which in turn constrains the robustness of CapDepth in challenging scenarios.
These results demonstrate that CapDepth maintains robustness under input texts of varying quality, and further confirm the effectiveness of the proposed \template.

\begin{table}[!t]
\caption{
Additional quantitative comparisons and computational cost analysis on NYUv2 \cite{nyu} and KITTI \cite{kitti}.
CapDepth achieves a favorable precision-FLOPs trade-off and outperforms Marigold in both accuracy and efficiency.
Notably, non-Lambertian surfaces and adverse weather conditions are rarely present in NYUv2 and KITTI, leading to a scarcity of visual ambiguities induced by these challenging factors.
}
\label{tab:flops}
\setlength{\tabcolsep}{0.5pt}
\begin{tabular}{cccccccc}
    \toprule
\multirow{2}{*}{Method}  & \multicolumn{2}{c}{NYUv2 \cite{nyu}}  & \multicolumn{2}{c}{KITTI \cite{kitti}}& FLOPs& Params& Latency \\
& AbsRel $\downarrow$ & $\delta_1 \uparrow$ & AbsRel $\downarrow$ & $\delta_1 \uparrow$& (G) $\downarrow$& (M) $\downarrow$&(ms) $\downarrow$ \\  
    \midrule
      Marigold \cite{marigold} & 5.5 & 96.4 & 9.9 & 91.6& 4237 & 949& 3180  \\
      VPD \cite{vpd} & 5.5 & 96.7 & 8.6 & 92.3& 978 & 991& 140  \\
      WorDepth \cite{wordepth}  & 8.8 & 93.1 & 10.8 & 88.4& 379 & 279& 79 \\
      Ours & {5.1} & {97.2} & {8.2} & {92.9}& 920 & 930& 342  \\ 
    \bottomrule
\end{tabular}
\end{table}

\begin{table}[!t]
\caption{
Ablation studies of the mask $M$ in the progressive masked querying blocks of the \encoder~in non-Lambertian surfaces (ToM surfaces on Booster \cite{booster}) and adverse weather conditions (rainy night on nuScenes \cite{nuscenes}).
}
\label{tab:ab_m}
\setlength{\tabcolsep}{6pt}
\begin{tabular}{ccccc}
    \toprule
\multirow{2}{*}{Method} & \multicolumn{2}{c}{ToM} & \multicolumn{2}{c}{night-rain} \\
& AbsRel$\downarrow$ & $\delta_1\uparrow$ & AbsRel$\downarrow$ & $\delta_1\uparrow$ \\
    \midrule
Random Binary Mask       & 4.1 & 98.1 & 29.7 & 64.5  \\
Ours & \textbf{3.9} & \textbf{98.7} & \textbf{24.9} & \textbf{68.3}  \\
    \bottomrule
    \end{tabular}
\end{table}

\begin{table}[!t]
\caption{
Ablation studies of the proposed \encoder~and \decoder~design in non-Lambertian surfaces (ToM surfaces on Booster \cite{booster}) and adverse weather conditions (rainy night on nuScenes \cite{nuscenes}).
(1) We replace the progressive masked querying blocks in the \encoder~ with the standard transformer layers.
(2) We replace the stable adaptive layer normalization (SAdaLN) in the \decoder~with cross attention.
}
\label{tab:others}
\setlength{\tabcolsep}{10pt}
\begin{tabular}{ccccc}
    \toprule
\multirow{2}{*}{Method} & \multicolumn{2}{c}{ToM} & \multicolumn{2}{c}{night-rain} \\
& AbsRel$\downarrow$ & $\delta_1\uparrow$ & AbsRel$\downarrow$ & $\delta_1\uparrow$ \\
    \midrule
(1)       & 4.9 & 97.5 & 29.9 & 63.1  \\
(2) & 4.2 & 97.3 & 30.2 & 63.4  \\
Ours & \textbf{3.9} & \textbf{98.7} & \textbf{24.9} & \textbf{68.3}  \\
    \bottomrule
    \end{tabular}
\end{table}

\subsection{Computational Cost}
We provide additional quantitative comparisons on NYUv2 \cite{nyu} and KITTI \cite{kitti} and computational cost analysis in Tab.~\ref{tab:flops}.
Results show that, compared with previous language-integrated methods (VPD \cite{vpd} and WorDepth \cite{wordepth}) and the method using the same diffusion U-Net backbone (Marigold \cite{marigold}), our method achieves a favorable precision-FLOPs trade-off and outperforms the image-only baseline (Marigold) in both accuracy and efficiency.

\subsection{The Mask $M$}
We conduct additional ablation studies for the mask $M$ in the \encoder.
Specifically, we replace $M$ with the random binary mask.
Results in Tab.~\ref{tab:ab_m} show that, when the mask is irregular, CapDepth receives misaligned language guidance, demonstrating the effectiveness of the proposed learnable soft mask $M$.

\subsection{Other Design Baselines}

We provide additional ablation studies of the proposed \encoder~and \decoder~design.
Specifically, we replace the progressive masked querying blocks in the \encoder~ with the standard transformer layers, and replace the stable adaptive layer normalization (SAdaLN) in the \decoder~with cross attention.
As shown in Tab.~\ref{tab:others}, results demonstrate that both the proposed \encoder~and \decoder~designs are important for improving robustness across non-Lambertian surfaces and adverse weather conditions.

\clearpage

\clearpage

\clearpage

\clearpage

\end{document}